\documentclass[a4paper,11pt,twocolumn]{article}
\usepackage[cmex10]{amsmath}
\usepackage{amssymb}
\usepackage{amsfonts}
\usepackage{amsthm}
\usepackage{algorithm}
\usepackage{algpseudocode}
\usepackage{graphicx}
\usepackage{dblfloatfix}
\usepackage{subcaption}
\usepackage{epsfig}
\usepackage{cite}
\usepackage{tensor}
\usepackage{comment}
\usepackage{color}
\definecolor{rephrase}{RGB}{0, 0, 0}
\definecolor{rephrase2}{RGB}{0, 0, 0}
\usepackage{diagbox}
\usepackage{gensymb}
\usepackage{environ}         
\usepackage{etoolbox}        

\usepackage{booktabs}
\usepackage{wasysym}
\usepackage{bm}

\usepackage[flushleft]{threeparttable}
 
\graphicspath{{figures/}}
\DeclareGraphicsExtensions{.eps}
\theoremstyle{definition}

\newlength{\myl}
\let\origequation=\equation
\let\origendequation=\endequation

\RenewEnviron{equation}{
  \settowidth{\myl}{$\BODY$}                       
  \origequation
  \ifdimcomp{\the\linewidth}{>}{\the\myl}
  {\ensuremath{\BODY}}                             
  {\resizebox{\linewidth}{!}{\ensuremath{\BODY}}}  
  \origendequation
}

\usepackage{lineno,hyperref}
\usepackage{authblk}
\usepackage[section]{placeins}
\usepackage{placeins}
\begin{document}

\title{Differential Mapping Spiking Neural Network for Sensor-Based Robot Control}

\author[a]{Omar Zahra}
\author[b]{Silvia Tolu}
\author[a]{David Navarro-Alarcon}
\affil[a]{The Hong Kong Polytechnic University, Hong Kong}
\affil[b]{Technical University of Denmark, Denmark}
\date{}


\maketitle

\begin{abstract}
In this work, a spiking neural network (SNN) is proposed for approximating differential sensorimotor maps of robotic systems. 
The computed model is used as a local Jacobian-like projection that relates changes in sensor space to changes in motor space. 
The SNN consists of an input (sensory) layer and an output (motor) layer connected through plastic synapses, with inter-inhibitory connections at the output layer. 
Spiking neurons are modeled as Izhikevich neurons with a synaptic learning rule based on spike timing-dependent plasticity. 
Feedback data from proprioceptive and exteroceptive sensors are encoded and fed into the input layer through a motor babbling process. 
As the main challenge to building an efficient SNN is to tune its parameters, we present an intuitive tuning method that considerably reduces the number of neurons and the amount of data required for training.
Our proposed architecture represents a biologically plausible neural controller that is capable of handling noisy sensor readings to guide robot movements in real-time. 
Experimental results are presented to validate the control methodology with a vision-guided robot.
\end{abstract}



\section{INTRODUCTION}
Sensor-guided object manipulation is one of the most fundamental tasks in robotics, with many possible approaches to perform it\cite{navarro2019model}. 
Conventional methods typically rely on mathematical modeling of the observed end-effector pose and its related joint configuration.
These methods provide accurate solutions, however, they require an exact knowledge of the analytical sensor-motor relations (which might not be known); furthermore, these conventional methods are generally not provided with adaptation mechanisms to cope with uncertainties/changes in the sensor setup.

Among the many interesting cognitive abilities of humans (and animals, in general) is the motor babbling process that leads to the formation of sensorimotor maps\cite{motor_babbling}. 
By relying on such adaptive maps, humans can learn to perform many motion tasks in an efficient way. 
These advanced capabilities have motivated many research studies that attempt to artificially replicate such skills in robots and machines \cite{aoki2016learning,xiong2015adaptive}.
Our aim in this work is to develop a bio-inspired adaptive computational method to guide the motion of robots with real-time sensory information and a limited amount of data.

Data-driven computational maps have been previously built for approximating unknown sensorimotor relations\cite{pinto2016supersizing,pierson2017deep,arulkumaran2017deep}.
One common limitation of these classical approaches is the demand for a high number of training data points and high computational power, which is impractical in many cases.

By drawing inspiration from the central nervous system, artificial neural networks (ANN) have been built and used for many decades. 
It started with the McCulloch Pitts model as the first generation of ANN by using binary computing units\cite{mcculloch1943logical}, followed by the second generation utilizing mainly the sigmoidal (or tanh) activation functions to make it more capable of approximating non-linear functions\cite{first_nn}. 
Taking one step further, spiking neuronal networks (SNN) (representing the third generation of ANN) are designed to incorporate most of the neuronal dynamics which provide them with more complex and realistic firing patterns based on spikes \cite{maass1997networks}.
In this work, we have built an adaptive SNN-based model to guide robots with sensory feedback.

The main exclusive property of SNN is the incorporation of a temporal dimension; note that the relative timing of spikes (and spikes sequences) enables the encoding of useful information. 
As in real biological systems, SNNs hold an advantage for real-time processing (as concluded in \cite{thorpe1996speed} for the visual system) and multiplexing of information (such as amplitude and frequency in the auditory system \cite{wang2005spiking}). \textcolor{rephrase}{For robotic applications, the SNN allows building computational models of brain regions to imitate intelligent behaviour in living organisms to a great extent, and in some cases even reveal the mysteries of the inner workings of the brain \cite{krichmar2018neurorobotics,luque2011adaptive,ojeda2017scalable}. The currently rising neuromorphic chips allow real-time operation of such models while conserving power greatly compared to conventional systems \cite{furber2016large}.}
While an SNN allows building a more biologically plausible system, its complexity makes it more difficult to predict and analyze the system behaviour. 
To this end, various methods can be used, e.g. simplifying the network's mathematical description \cite{schoner2016dynamic} or applying techniques for tuning the parameters to achieve the desired performance \cite{hbp_hand}. 
The latter approach is the one addressed in this work.

Several studies have provided examples of the application of SNN in robotics \cite{bing2018survey}.
However, in most studies, robots were only controlled in a simulation environment\cite{corchado2019integration}. 
The main reason is due to the large size of the neural networks used in these works, which makes it impractical for real-time operations.
In \cite{hbp_hand}, a cognitive architecture is used for controlling a robotic hand to perform grasping motions. 
In \cite{tieck2018controlling}, an SNN was used for learning motion primitives of a robot so to move in three axes (left-right, up-down and far-near). 
In \cite{narayan2012self}, a self-organizing architecture was used to build an SNN representing spatio-motor transformations (i.e. kinematics) of a two degrees of freedom (DOF) robot. 
Despite its good performance and biologically plausibility, the network's size and limited scalability make it impractical for real-time control.

In \cite{wu20082d}, an SNN with learning based on spike timing-dependent plasticity (STDP) was used to build the kinematics of a robot. 
Although the synaptic connections illustrate the ability to approximate the kinematic relation, the approximation error is not evident. \textcolor{rephrase}{Additionally, an intermediate layer is required to scale up the dimensions of the input sensory data, and consequently, the relative computational power.}

In this paper, we propose an SNN-based control architecture to guide robots with sensor feedback.
Our proposed neural controller adaptively builds the differential map that correlates end-effector velocities (as measured by an external vision sensor) with joint angular velocities. 
In other words, it effectively works as a local Jacobian-like transformation between different (sensor-motor) spaces. 
This new method is characterized by inter-inhibitory connections at the output layer, real-time operation and a fewer number of neurons, compared to previous works in literature \textcolor{rephrase}{\cite{hbp_hand,tieck2018controlling,murray}. The inter-inhibitory connections distinguish the proposed architecture from that introduced in \cite{murray}.} 

\textcolor{rephrase2}{These inter-inhibitory connections with a winner-take-all (WTA) effect is studied as well from a probabilistic perspective in \cite{mass_stdp_wta}. Such effect is proven to be an essential component of a special setup, that allows the emergence of an approximation of a forward sampler \cite{koller2009probabilistic} in a Hidden Markov Model (HMM) \cite{rabiner1989tutorial,murty2011hidden} capable of online learning. Moreover, the study in \cite{murray} suggests to divide the input into separate bins to improve the learning but both the formulation and the explanation of how this is achieved is lacking, and the method to set and tune the network parameters is not mentioned.}
To the best of the authors' knowledge, this is the first study to report an SNN-based method capable of forming the sensor-motor differential map in a computationally efficient way (resulting from an intuitive guideline to adjust its parameters). \textcolor{rephrase}{Additionally, the network relies on biologically plausible models of neurons and synapses, which are more complex compared to other relatively simpler models, to both preserve features of the biological nervous system as much as possible and serve as a future building block for simulating the whole hierarchy of the biological motor system. An example of this is the simple Izhikevich neuron model \cite{izhikevich2003simple} used in the proposed network, instead of the commonly used Leaky Integrate and Fire model (LIF)\cite{abbott1999lapicque}.}
To validate the theory, we present a detailed experimental study with a robotic platform performing vision-guided manipulation tasks.

The rest of this paper is structured as follows: Sec. II describes the developed spiking neural network; Sec. III presents the verification of the method proposed through a dummy test; Sec. IV presents the experimental results; Sec. V discusses the methods and results; Sec. VI gives the conclusions.

\section{METHODS}\label{sec:methods}

Many studies have suggested that humans use internal models to represent perception and action \cite{wolpert1998multiple,blakemore2000can,maravita2004tools}. 
Some researchers have built computational models of brain areas (e.g the cerebellum) responsible for the generation and coordination of fine motor actions \cite{abadia2019robot,vannucci2017comprehensive}.
\textcolor{rephrase}{Unlike in traditional robot control (where sensory transformations to motion commands are solved analytically), in the human brain (motor cortex) the sensorimotor relations are encoded by specific neural circuits.}
To carry out a typical (eye-to-hand) visual servoing task, we must first establish the kinematic relation between the joint velocities and the measured end-effector motion \cite{Journals:Chaumette2006}.
This model can be formulated as $\Dot{x}=J(\theta)\Dot{\theta}$, such that:
\begin{equation}
J(\theta) = 
\begin{bmatrix}
\frac{\partial x_{1}}{\partial \theta_{1}} & \cdots & \frac{\partial x_{1}}{\partial \theta_{m}} \\ 
\vdots & \ddots & \vdots\\
\frac{\partial x_{n}}{\partial \theta_{1}} & \cdots & \frac{\partial x_{n}}{\partial \theta_{m}} 
\end{bmatrix}
\end{equation}
where $\Dot{x}\in\mathbb R^n$, $\theta\in\mathbb R^m$, $\Dot{\theta}\in\mathbb R^m$ and $J(\theta)\in\mathbb R^{n\times m}$ are the (observed) spatial velocity, joint angles, joint velocities and the Jacobian matrix, respectively (without loss of generality, we assume that $m\ge n$). 
For velocity-based control it is necessary to estimate the joint velocities to achieve a certain spatial velocity. 
This expression is denoted as:
\begin{equation}
\Dot{\theta} = J^{\#}(\theta) \Dot{x}
\label{eq:inv_j}
\end{equation}
where $J^{\#}$ is the pseudo-inverse of the $J$.
In this work, a differential mapping spiking neural network (DMSNN) is proposed to build a computational model analogous to the Jacobian transformation relating changes in 1 joint-space DoF to 1 task-space DoF.

\begin{figure*}[t]
\centering
\includegraphics[width=0.78\linewidth]{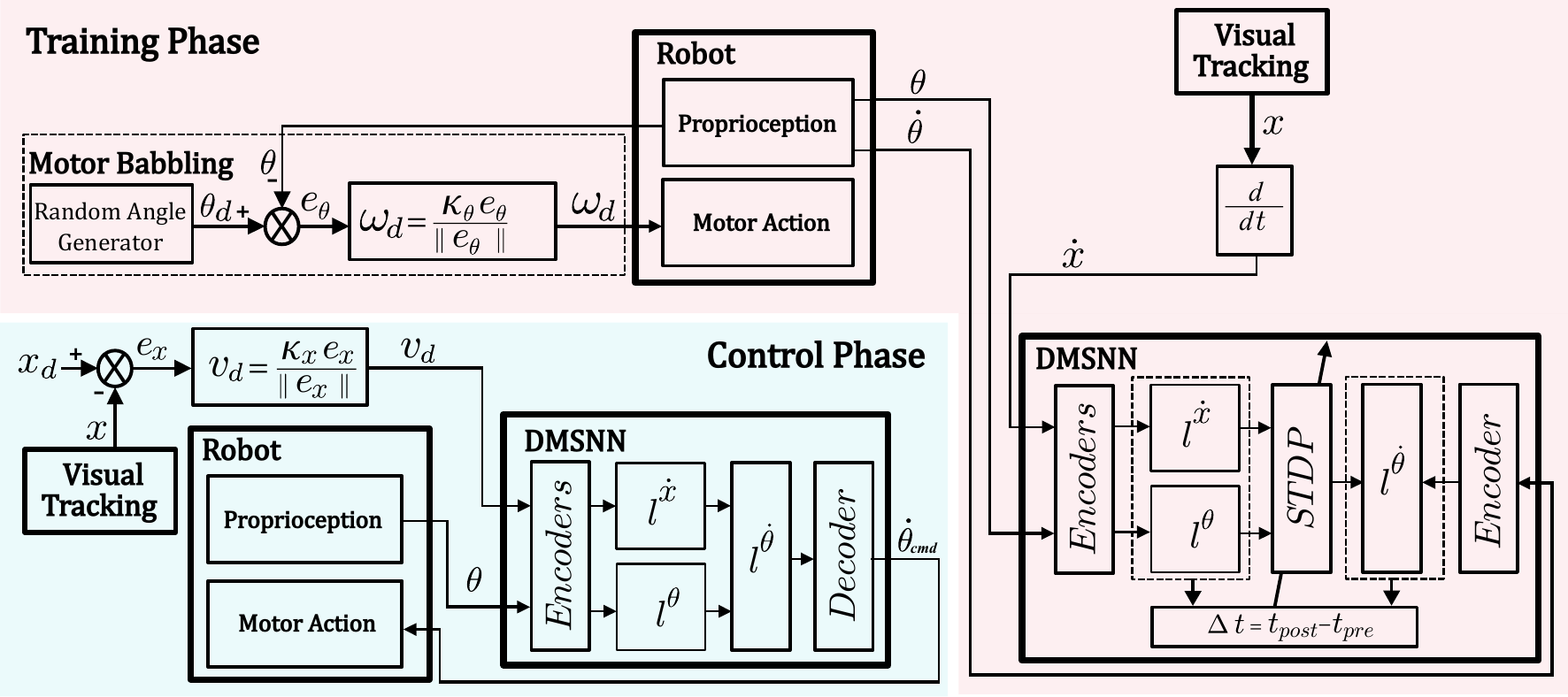} 
\caption{A schematic diagram for both the training and control phases for the DMSNN. The signals introduced to each neuron bundle are depicted. During the training phase, the robot motion is guided by motor babbling in joint space. During the control phase, the robot motion is guided by motor commands decoded from the activity of motor neurons at the output layer.}
\label{fig:train_test}
\end{figure*}

As shown in Fig. \ref{fig:train_test}, the network is first subject to the training phase in which both sensory readings and motor commands are fed to the network through motor babbling by executing random motions. 
After training the network for several iterations, the differential mapping is formed by the modulation of the connection between the input and output layers. 
Then, the network can be used during the control phase to guide the robot through the estimation of the required motor command to reach the desired target by feeding the sensory information only to the input layer. \textcolor{rephrase}{This can be seen as an approximation of the motor cortex which is responsible for converting the desired motion into a motor command to be executed by other regions in the brain \cite{kalaska2009intention}.}
Details of the proposed network are described in the rest of this section.

\begin{figure}[!b]
	\centering
	\includegraphics[width=\columnwidth]{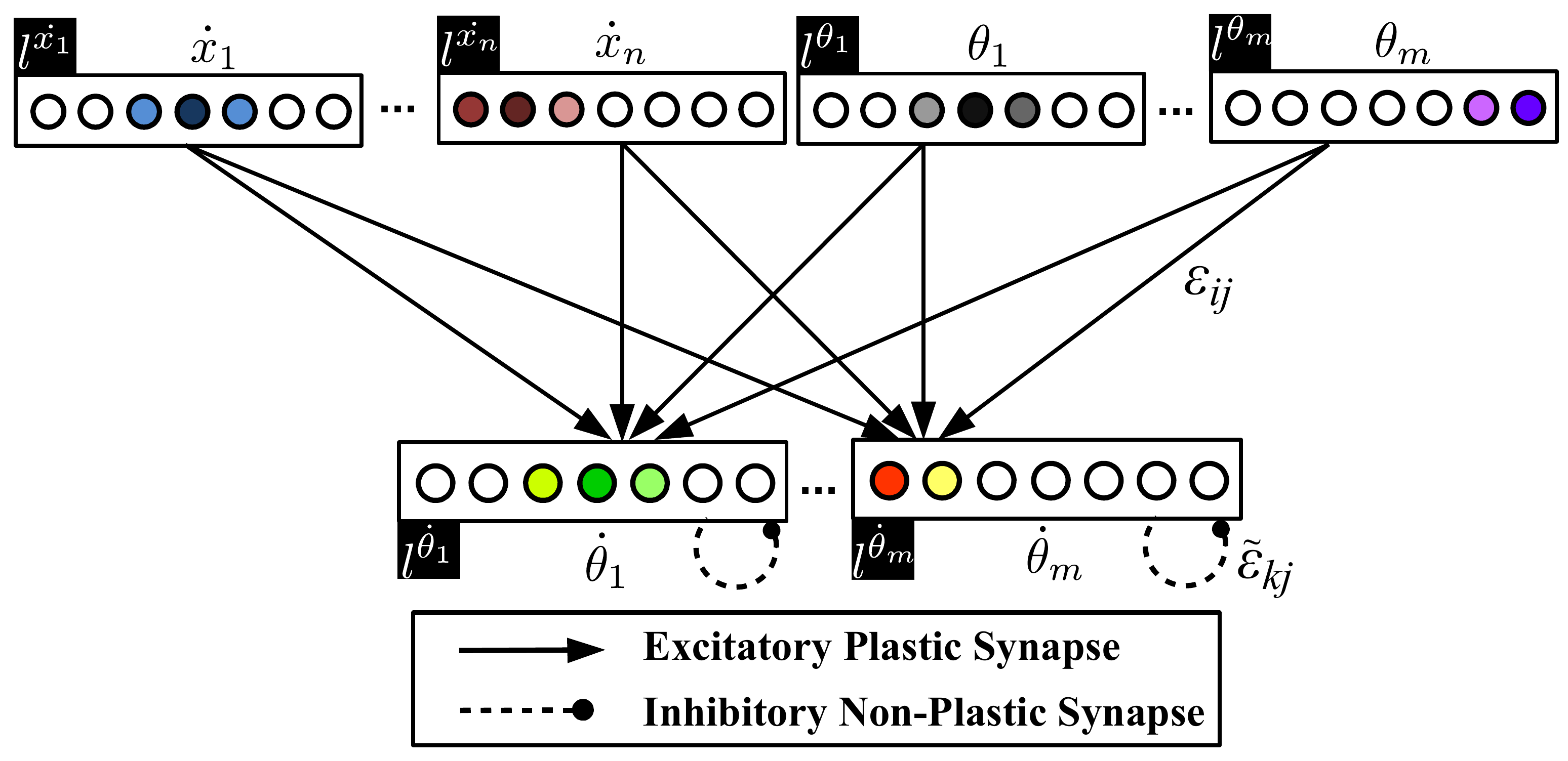}
	\caption{The layout of the proposed spiking neural network. Input (sensory) neurons are connected to output (motor) neurons through excitatory and inhibitory plastic synapses ${\varepsilon}_{ij}$; neurons in each motor bundle are interconnected through inhibitory non-plastic synapses $\tilde{\varepsilon}_{ij}$.}
\label{fig:net}
\end{figure}

\subsection{Network Layout}
The network layout of the proposed neural controller is shown in Fig. \ref{fig:net}, where each dimension of sensory input (joint angles and spatial velocity) and motor output (joint velocity) is represented by a one-dimensional array (bundle) of neurons. 
The complete network consists of $n+2m$ bundles of neurons. 
The input sensory layer consists of bundles $l^{\theta_{1:m}}$ encoding the joint angles and $l^{\Dot{x}_{1:n}}$ encoding the spatial velocity, while the output motor layer consists of $l^{\Dot{\theta}_{1:m}}$ encoding the joint velocity controls.
Each sensory neuron is connected through an excitatory and inhibitory synapse (which is omitted in Fig. \ref{fig:net} for clarity) to each motor neuron. 
This acts as a substitute for adding an inhibitory interneuron and allows for stable learning dynamics while avoiding an unbounded increase in connection strength and neuron activity. 
\textcolor{rephrase}{Additionally, each of the neurons encoding the output at each bundle (dimension) $\Dot{\theta}_{1:m}$ is connected to the neurons of the same bundle through non-plastic inhibitory connections:
\begin{equation}
\label{eq:inter_inhib}
\tilde{\varepsilon}_{kj} = \exp \left( \dfrac{-(k-j)^{2}}{(\sigma_{n} N_{l})^{2}} \right)-1
\end{equation}
where $\tilde{\varepsilon}_{kj}$ denotes the strength of the \emph{non-plastic} connection between neurons $k$ and $j$, $\sigma_{n}$ is the standard deviation, and $N_{l}$ is the number of neurons in the bundle.}
Therefore, the further the neuron is, the stronger the inhibition activity is. 
This approximates the behavior of a winner-take-all effect (WTA), but with the change in the inhibitory value depending on the proximity to the winner neuron (ensuring a continuous and more robust output).

\subsection{Training Phase and Control Phase}

For the proposed network to form the desired differential map, the information needs to be input/encoded into the network and extracted/decoded in a proper way.
The input to the sensory layers (during training and control phases) and motor layers (during the training phase only) is calculated for each neuron based on its preferred (central) value $\psi_{c}$. 
Thus, this Gaussian distribution models the network's tuning curve. 
The firing rate for a certain input can be formulated as:
\begin{equation}
\alpha_{i}(t) = \exp {\left(\dfrac{-\Vert{\psi -\psi_{c}  \Vert^{2}}}{2\sigma^{2}}\right)}
\end{equation}
where $\psi$ is the input value, and $\sigma$ is calculated based on the number of neurons per layer $N_{l}$, and the range of change of the variable to be encoded from $\Psi_{min}$ to $\Psi_{max}$.
This leads to the contribution of the whole layer to encode a particular value (a process that can be interpreted as ``population coding'' \cite{amari2003handbook}). 

To get the estimated output from the network, a proper decoding function has to be chosen. 
Among the various decoding methods, the central neuron voting scheme is selected to calculate the decoded value corresponding to the firing rate in all the neurons of a layer.
This can be modeled, for a specific time window, as follows:
\begin{equation}
\psi_{est}= \dfrac{\Sigma \psi_{i}.\alpha_{i}}{\Sigma \alpha_{i}}
\end{equation}
where $\psi_{i}$ is the central value of neuron $i$ in the bundle $l^{\Psi}$, and $\psi_{est}$ is the estimated (decoded) value of the output\cite{amari2003handbook}.

Fig. \ref{fig:train_test} depicts a schematic diagram of the proposed SNN-based method.
Firstly, the motor babbling process initiates the training phases by providing
the motor commands (joint velocity command ($\omega_{d}$) for the robot
to move linearly in the joint space through numerous random targets, as it is formulated in:

\begin{equation}
    \omega_{d} = \kappa_{\theta} \frac{e_{\theta}}{\|e_{\theta}\|}
\end{equation}

where $e_{\theta} = \theta_{d}-\theta$ is the error between the randomly generated desired joint angles ($\theta_{d}$) and the current joint angles ($\theta$).

This joint velocity command is scaled by the gain $\kappa_{\theta}>0$, which is also varied randomly, within a certain range, during the babbling, to generate richer training data. 
During babbling motions, sensory information and motor commands are fed into the network to guide the modulation of the plastic synapses between the input and output layers through STDP. 
Sensory information is introduced from proprioception ($\theta$ and $\dot{\theta}$) and the external sensor (the visual velocity $\dot{x}$), which are then encoded. 
The difference in time of spikes generated in sensory and motor neurons $\Delta t = t_{post} - t_ {pre}$ controls the amount of change in the synapses' strength. 
After a sufficient number of iterations (decided depending on the size of work space, learning rate and desired precision as explained is subsection \ref{subsection:tuning_params} , the robot is ready to perform a sensor-guided motion task. 

\begin{equation}
    v_{d} = \kappa_x \frac{e_x}{\|e_x\|} 
\end{equation}
where $e_x = x_d - x$ denotes the feedback spatial error and $\kappa_x>0$ a variable gain that regulates the velocity by which the robot is driven towards the target $x_d$ from the current position $x$.
Finally, the motor command $\Dot{\theta}_{cmd}$ is decoded from the spikes generated at the output bundles $l^{\Dot{\theta}_{1:m}}$ as follows:
\begin{equation}
\dot{\theta}_{cmd} = \psi_{est} 
\label{eq:decode}
\end{equation}
This value is then fed into the robot servo controller to guide its motion towards $x_d$.

\subsection{Neuron Model}
Among the models available for spiking neurons is the Hodgkin and Huxley model that explains the mechanism of triggering an action potential and its propagation \cite{hodgkin1952quantitative}. 
The mathematical model is composed of a set of non-linear differential equations, which makes it computationally intense. 
However, this model is the most biologically plausible among all available models.

Another model that is widely used is the Leaky Integrate and Fire model (LIF). 
In this model, the behavior of a neuron is approximated as a simple RC circuit with a low-pass filter and a switch with a thresholding effect \cite{abbott1999lapicque}. 
The RC circuit is first charged by an input current, which makes the voltage at the capacitor to increase until it reaches the threshold value. 
The switch opens and lets a pulse (spike) to be generated; the capacitor then starts to build up again. 
This model is simple and has a low computational cost, however, compared to Hodgkin-Huxley's model, it is not very biologically plausible. 

In 2003, Izhikevich developed a model that can reproduce the various firing patterns recorded by different neurons in different brain regions\cite{izhikevich2003simple}. 
This model is chosen for our study as it holds a balance between a reasonable computational cost while preserving the biological plausibility.
The model can be described by the following set of differential equations:
\begin{align}
\dot{v} & = f(v,u) = 0.04v^2+5v+140-u+I    \label{eq:update_v} \\
\dot{u} & = g(v,u) = a(bv-u)               \label{eq:update_u}
\end{align}
After a spike occurs, the membrane potential is reset as:
\begin{equation}
\label{eq:reset}
\text{if }v\ge30 \text{ mV}, \quad
\text{then } v\leftarrow c,~ u\leftarrow (u+d)
\end{equation}
where $v$ is the membrane potential and $u$ is the membrane recovery variable as shown in Fig.\ref{fig:phase_portrait}.
The parameter $a$ determines the time constant for recovery, $b$ determines the sensitivity to fluctuations below the threshold value, 
$c$ gives the value of the membrane potential after a spike is triggered, 
and $d$ gives the value of the recovery variable after a spike is triggered. 
The term $I$ represents the summation of external currents introduced.

\subsection{Synaptic Connections}
Synapses are the connections that transmit signals between two neurons. Let us denote by ${\varepsilon}_{ij}$ the weight/strength of the synapse.
The transmission acts only in one direction, such that signals are carried from the $i$th presynaptic to the $j$th postsynaptic neuron. 
The information transmitted through synapses is usually encoded in the form of spikes (or action potentials). 
Different theories have been presented for the way of encoding and decoding such information in our brains\cite{amari2003handbook}.
The synaptic connections can either be excitatory or inhibitory, and plastic or non-plastic. 
The excitatory synapses are the connections that are more likely to increase the activity of post-synaptic neurons with the increase in the activity of the presynaptic neuron, while the inhibitory synapses decrease that likelihood.
The synapses connecting the input layer to the output layer are plastic (which means that its strength is subject to change).
Let us denote by $\Delta {\varepsilon}_{ij}[t]$ the synapse's change of strength at the time instance $t$, which satisfies the following discrete update rule:
\begin{equation}
    \varepsilon_{ij}[t+1] = \varepsilon_{ij}[t] + \Delta {\varepsilon}_{ij}[t]
\end{equation}

One of the very first learning rules to update the synaptic weights is the Hebbian Learning rule\cite{hebb2002organization}, whose basic formulation is:
\begin{equation}
    \Delta {\varepsilon}_{ij} =\eta a_{i}a_{j}
\end{equation}
where the scalar $\eta$ is the learning rate, $a_{i}$ and $a_{j}$ are the activities (or average firing rates) of the pre and postsynaptic neurons, respectively.
This rule strengthens the connections between strongly correlated variables, and has been shown to perform principal component analysis (PCA) \cite{oja1982simplified}. 
However, this type of learning does not take into consideration the \emph{time difference} between spikes, which is the main feature of SNNs.

Another learning rule that is more appealing from a biological perspective is STDP, where potentiation (increase) or depression (decrease) in the strength of the connections is dependent upon the relative timing of spikes that occur in presynaptic and postsynaptic neurons \cite{stdp}.
STDP is considered to be one temporal form of Hebbian learning \cite{caporale2008spike}, e.g. in \cite{gilson2012spectral}, it is shown that it can perform `kernel spectral component analysis' (kSCA), which resembles PCA. 
This attribute makes it suitable for mapping two spaces while successfully updating the synaptic weights.

\begin{figure}[!b]
\centering
\begin{subfigure}{0.5\columnwidth}
\centering
\includegraphics[width=0.95\linewidth]{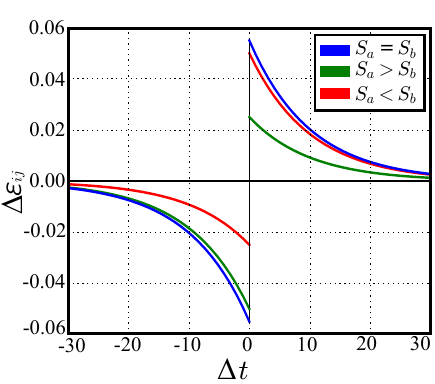} 
\caption{}
\label{fig:stdp_asym}
\end{subfigure}%
\begin{subfigure}{0.5\columnwidth}
\centering
\includegraphics[width=0.95\linewidth]{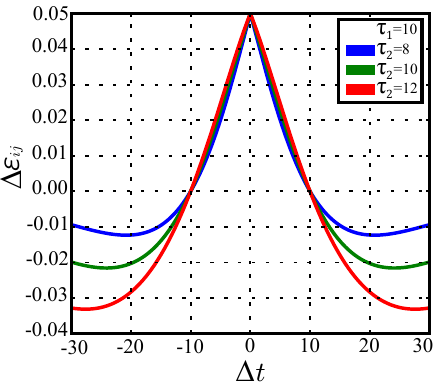}
\caption{}
\label{fig:stdp_sym}
\end{subfigure}
\caption{(a) Asymmetric and (b) Symmetric learning STDP rules. For asymmetric learning, $S_a$ and $S_b$ are plotted at different values while keeping $\tau_{a}$ and $\tau_{b}$ constant. For the symmetric one, $\tau_{2}$ is plotted at different values while keeping $\tau_{1}$ constant.}
\label{fig:stdp}
\end{figure} 

In the literature \cite{cutsuridis20082+}, two common patterns for STDP is either as symmetric \cite{woodin2003coincident} or antisymmetric \cite{stdp_asym_pic}, as depicted in Fig. \ref{fig:stdp}.
The antisymmetric model can be formulated as:
\begin{equation}
\Delta {\varepsilon}_{ij} = \left\{
        \begin{array}{ll}
            -S_{a} \exp \left( {-\Delta t}/{\tau_{a}} \right) & \quad \Delta t \leq 0 \\
            \\
             S_{b} \exp \left( {-\Delta t}/{\tau_{b}} \right) & \quad \Delta t > 0
        \end{array}
    \right.  
\label{eq:asym_STDP}
\end{equation}
where $S_{a}$ and $S_{b}$ are coefficients that control the magnitude of the synaptic depression and potentiation, respectively, where $\tau_{a}$ and $\tau_{b}$ determine the time window through which depression and potentiation occur. 
The asymmetric STDP is thus suitable for a learning process whenever the sequence of signals matters, e.g. if a spike arrives from the presynaptic neuron before the spike from a postsynaptic neuron (which results in a positive value for $\Delta t$, and thus the synapse's weight is potentiated).

\textcolor{rephrase}{The symmetric learning model is the most appropriate in this case given the continuous firing at both input and output layers. Furthermore, a symmetric STDP rule (with different reward modulated versions) was reported to be observed in hippocampus and prefrontal cortex in several studies \cite{ruan2014dopamine,zhang2009gain,brzosko2015retroactive} and was studied in \cite{hao2020biologically}. The symmetric STDP rule can be described by:}
\begin{equation}
\Delta {\varepsilon}_{ij} = S \left( 1-\left( \frac{\Delta t}{\tau_{1}} \right)^2\right) \exp\left(\frac{|\Delta t|}{\tau_{2} } \right)
\label{eq:sym_STDP}
\end{equation}
where $S$ is a coefficient that controls the magnitude of the synaptic change, the ratio between $\tau_{1}$ and $\tau_{2}$ decides the time window through which potentiation and depression occurs, and $\Delta t$ is the difference between the timing of spikes at post $t_{post}$ and pre $t_{pre}$ synaptic neurons. 
In this study, we use $S = 0.05$, $\tau_{1}=20$ms and $\tau_{2}=18$ms. 
In this learning model, the change in synapse's weight is controlled by the absolute value of $\Delta t$, but not the sign, i.e. the sequence of firing. 
The chosen time window for the pre and postsynaptic neurons for STDP is 30ms, which means as long as $-30 \leq \Delta t \leq 30$, it will still contribute to the modification of the synaptic strength.

\textcolor{rephrase}{With the absence of the hidden layer(s), non-linearity in the neuronal units, as well as the features of STDP, make it possible to learn the differential map. 
Such approach is supported by previous studies (see \cite{murray} and \cite{hbp_hand}), as illustrated in section \ref{sec:results}.}

\textcolor{rephrase2}{Additionally, the inter-inhibitory connections establish the WTA effect along with the STDP and the slight contribution from adjacent neurons (approximating the effect of lateral excitatory connections) develop an approximation of a Hidden Markov Model (HMM) online learning as described in \cite{mass_stdp_wta}. For a Markov process, the current state depends only on the current state and independent of the past states \cite{gagniuc2017markov}. This property is useful in our case as it guarantees the continuity of the output and avoiding jerky motions, but still capable of changing the output based on the current state in case of inaccurate estimations. Moreover, the HMM developed in that case proven to perform \textit{forward sampling}\cite{koller2009probabilistic} in the WTA circuit, and achieve an online stochastic approximation to \textit{expectation-maximization (EM)} parameter learning thanks to the STDP. So, from a probabilistic perspective, connections from the input layer provide the \textit{observations} for the HMM, while the WTA circuit provides an approximation for the \textit{E-step}, based on forward sampling. To study the proposed architecture and compare it to that in \cite{mass_stdp_wta}, it needs an extensive separate study to fully analyze the DMSNN from a statistical perspective.} 

\textcolor{rephrase}{To perform the vision-guided motion task with the proposed network, a careful setting of its parameters is needed, as explained in the next section.}

\subsection{Tuning the Network Parameters}
\label{subsection:tuning_params}
Tuning the various parameters for the neurons and synapses is a difficult task. 
Some basic rules can be used to guide the trial and error approach to choose a set of appropriate values. 
For example, as $u$ is the membrane recovery variable, it is responsible for the delivery of negative feedback to $v$, such that it resists change of the value of $v$.
The network's parameters $a, b, c$ and $d$ tune how $v$ and $u$ change and interact together over time. 

\begin{figure}[!b]
\centering

\begin{subfigure}{0.45\columnwidth}
\centering
\includegraphics[width=0.9\linewidth]{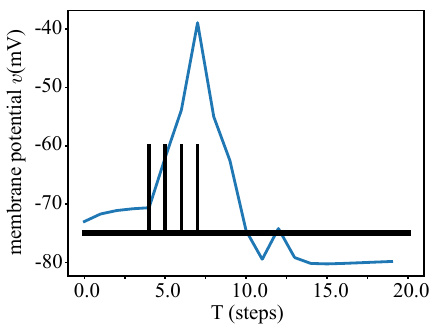} 
\caption{}
\label{fig:four_pre}
\end{subfigure}%
\begin{subfigure}{0.53\columnwidth}
\centering
\includegraphics[width=\linewidth]{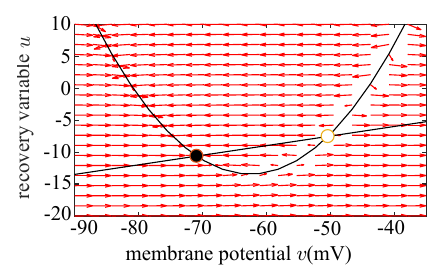}
\caption{}
\label{fig:phase_portrait}
\end{subfigure}
\caption{(a) Computational and (b) analytical analysis of a spiking neuron. In (a) an integrator neuron receives spikes from four neurons to cross a threshold value and trigger a spike, While in (b) a phase portrait of a spiking neuron is plotted.}
\label{fig:analysis_sn}
\end{figure} 

Fig. \ref{fig:phase_portrait} shows the phase portrait of the output neurons, as it plots the recovery potential versus the membrane potential. The two black curves represent the \textit{nullclines} for both $v$ and $u$, i.e. the line along which partial derivative equals zero which separates the planes of variation of $v$ and $u$. 
To obtain these nullclines and analyze the system, equations \ref{eq:update_v} and \ref{eq:update_u} are set equal to zero to obtain lines along which there is no change in $v$ and $u$, respectively.
The intersection of these nullclines forms attractor (stable) points or repeller (unstable) points \cite{Izhikevich_book}. 
From the location of attractors and repellers, the stability regions can be concluded.
The equilibrium points ($v^{*}$, $u^{*}$) are obtained by solving the equations of the parabola and line obtained from equating $f(v,u)$ and $g(v,u)$ together. 
The stability of these points is determined by the eigenvalues of the Linearization matrix $L$:
\begin{equation}
\label{eq:linearization}
L(v^*,u^*) = \begin{bmatrix} \frac{\partial f}{\partial v}(v^*,u^*) & \frac{\partial f}{\partial u}(v^*,u^*) \\ \frac{\partial g}{\partial v}(v^*,u^*) & \frac{\partial g}{\partial u}(v^*,u^*) \end{bmatrix} = 
\begin{bmatrix} 
0.08v^*+5 & -1 \\ ab & -a
\end{bmatrix}
\end{equation}

As the $v$-nullcline shifts upward, the attractor and repeller annihilate each other and merge into a saddle; any further upward shift leads to the disappearance of the saddle point. 

\begin{algorithm}[!b]
    \caption{Network Parameters Tuning}\label{alg1}
    \textbf{Input} \\
    \hspace*{\algorithmicindent} {$n$,$m$} = Number of dimensions of spaces to be encoded\\
    \hspace*{\algorithmicindent} $N_{l}$ = Number of neurons to encode each dimension\\
    \hspace*{\algorithmicindent} $Itr^{*}$ = Number of iterations to build the map\\
    \textbf{Output} \\
    \hspace*{\algorithmicindent} $a, b, c ,d$ = Neurons parameters\\
    \hspace*{\algorithmicindent} $I^{*}$ = Minimum input to have continuous spikes\\
    \hspace*{\algorithmicindent} $C_{I}, C_{E}$ = Maximum strength of Inhibitory and Excitatory connections, respectively\\
    \hspace*{\algorithmicindent} $S$ = Learning rate for STDP based synaptic connections\\
    \textbf{Routine}
    \begin{algorithmic}[1]
        \State {Assign initial values for neuron parameters $a, b, c ,d$}
        \While{$\xi \ge e_{th}$}
            \State Build the neuron model $f(v,u)$ and $g(v,u)$
            \State Solve for nullclines at $f(v,u)=0$ and $g(v,u)=0$
            \State Get $v^{*}$ and $u^{*}$ where $f(v^{*},u^{*})=g(v^{*},u^{*})=0$
            \State Evaluate $L(v^{*},u^{*})$ to check stability
            \State Obtain $I^{*}$ and hence $C_{E}$
            \State Set $S$ for given $Itr^{*}$
            \State Update values for $a, b, c ,d$
        \EndWhile
    \end{algorithmic}
\end{algorithm}

As shown in Algorithm \ref{alg1}, to use an SNN to model a certain system, initial values should be assigned for neuron parameters $a, b, c ,d$ based on their role within the neuronal layer. 
In this work, the motor neurons are modeled as \textit{integrator neurons} as it acts as a coincidence detector, such that it triggers a spike by accumulating closely timed signals as illustrated in Fig. \ref{fig:four_pre}. The initial values for the neuron parameters are $a=0.02$, $b=-0.1$, $c=-55$, and $d=6$. 
The sensory neurons are modeled as \textit{fast spiking} neurons providing high frequency spikes and initialized with $a=0.1$, $b=0.2$, $c=-65$, and $d=2$. 
The above-mentioned initial values are suggested in \cite{izhikevich2004model}.

The parameters' variation produces specific properties in the system, which can be used as a guideline to adjust the neuron's firing pattern:
(i) The value of $a$ controls the decay rate of $u$. 
This can be clearly noticed when comparing the firing behavior of low-threshold spiking (LTS) neurons with resonator (RZ) neurons. RZ neurons have a higher value of a (to set sub-threshold oscillations and resonate at a narrow band of frequencies) than LTS neurons, but same values of b,c and d. 
(ii) The parameter $b$ controls the sensitivity of $u$ to changes in the membrane potential $v$ below the threshold value. 
The integrator neuron has a low value of $b$ which is why many spikes with small time interval in between are needed to trigger a spike. 
(iii) The parameter $c$ describes the value to which $v$ is reset after firing. 
Therefore, decreasing its value enables to create spikes bursts since it makes the recovery to the original threshold value faster. 
(iv) The parameter $d$ describes the reset in the value of $u$ after a spike occurs, thus, lowering it creates a higher firing frequency for the same input current.

After setting the neuron parameters, the value for $I^{*}$ and $C_{E}$ can be concluded by analyzing the stability at $v^{*}$ and $u^{*}$ such that a specific firing rate at the output neurons is obtained for a certain input from multiple input neurons. 
$I^{*}$ is even more critical to estimate in our study, as selective disinhibition has to be achieved and the neuron has to be maintained at the verge of firing to avoid excessive firing \cite{sridharan2015selective}.
This ensures that only the correct motor neurons fire upon excitation of the corresponding sensory neurons. 

To obtain $I^{*}$, we first equate $g(v^{*},u^{*})=a(bv^{*}-u^{*})=0$, and solve for $u^{*}=bv^{*}$. 
Then, substitute $u^{*}$ into $f(v^{*},u^{*})=0$ such that $ 0.04v^{*2}+(5-b)v^{*}+140+I=0$. The intersection of the $u$ and $v$ nullclines at one point is when the neuron starts to give continuous spikes, which means there is only one solution for the quadratic equation. 
The equilibrium points are given by $v^{*}=-(5-b)/(2\times0.04)$ and $u^{*}=bv^{*}$. 
The value of $I^{*}$ is finally computed by solving $f(v^{*},u^{*})=0$.
The parameter $C_{E}$ must be chosen such that at the end of the training phase the selected firing behavior is still maintained.
After setting $C_{E}$, the spiking behavior is tested. 
The chosen values for the motor neurons must not allow evoking spikes at low spiking frequency from sensory neurons. 
Note that increasing the frequency makes the learning process more susceptible to noise. 
Therefore, the motor neuron parameters are to be modified instead. 
The parameter $b$ is then incremented slightly until a satisfactory performance is obtained.

\begin{figure}[!b]
\centering
\includegraphics[width=\linewidth]{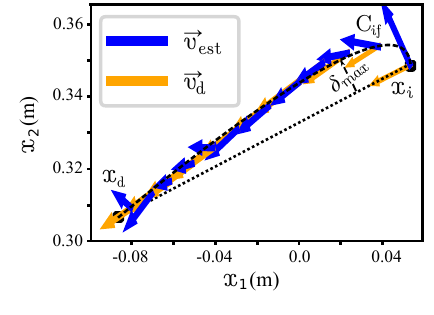} 
\caption{A planar robot driven by the DMSNN, initially at $x_{i}$ reaching to a target at $x_{d}$ and at each time step both the actual velocity ($\Vec{v}_{est}$) and the desired velocity vector ($\Vec{v}_{d}$) are calculated.}
\label{fig:servo_2d}
\end{figure} 

Depending on the size of the data set and the number of neurons in each bundle, the number of iterations $Itr^{*}$ must be defined to build the map, which in turn determines the value of the learning rate gain $S$. 
If $Itr^{*}$ is relatively small, it will lead to a large $S$, which often leads to instability and noise sensitivity. 
A large $Itr^{*}$ results in an exhaustive (computationally demanding) training process.

To quantify the accuracy of the computed differential map approximated by the network, we define the following metric for the accuracy of estimations as introduced in Algorithm \ref{alg1}:
\begin{equation}
\xi=\frac{1}{N}\sum_{n=1}^{N}\sqrt{(\Vec{v}_{d}-\Vec{v}_{est})^{2}}
\end{equation}
which is simply the difference between the desired spatial velocity $\Vec{v}_{d}$ and the spatial velocity $\Vec{v}_{est}$ obtained upon execution of the estimated motor command $\dot{\theta}_{cmd}$ as shown in Fig. \ref{fig:servo_2d} (calculated from the decoding equation \ref{eq:decode}); the scalar $N>0$ is the number of points in the workspace over which the difference is measured to obtain a mean error value. 
The tuning process continues until the value of $\xi$ is below some threshold value $e_{th}$, indicating that the network's performance is acceptable. The defined metric can be used for future improvements of the network by searching for the optimal network parameters using evolutionary algorithms \cite{carlson2014efficient}, or building a confidence map to guide the babbling process \cite{Saegusa2009}.

\section{SIMULATION RESULTS}\label{sec:method_ver}
\subsection{Network Layout Selection}\label{subsection:summation}
\begin{figure}[b]
\centering
\begin{subfigure}{0.4\columnwidth}
\centering
\includegraphics[width=\linewidth]{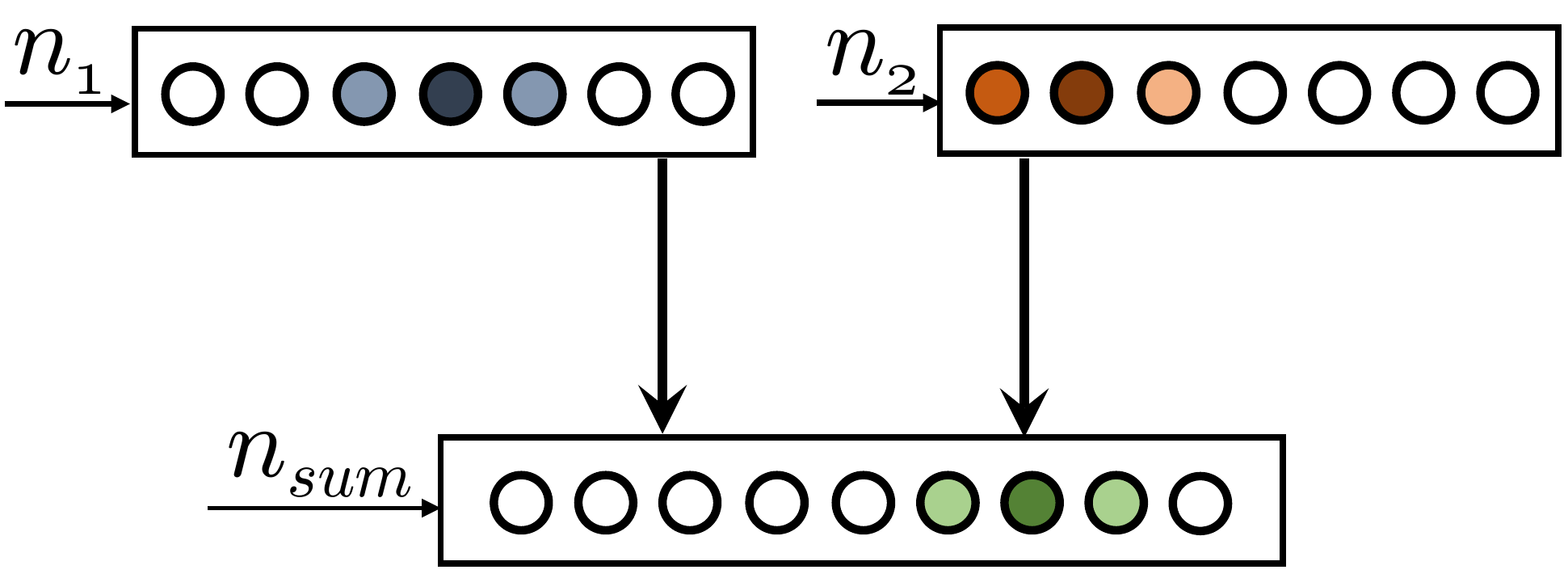} 
\caption{Separate encoding}
\label{fig:sum1d}
\end{subfigure}%
\begin{subfigure}{0.4\columnwidth}
\centering
\includegraphics[width=0.8\linewidth]{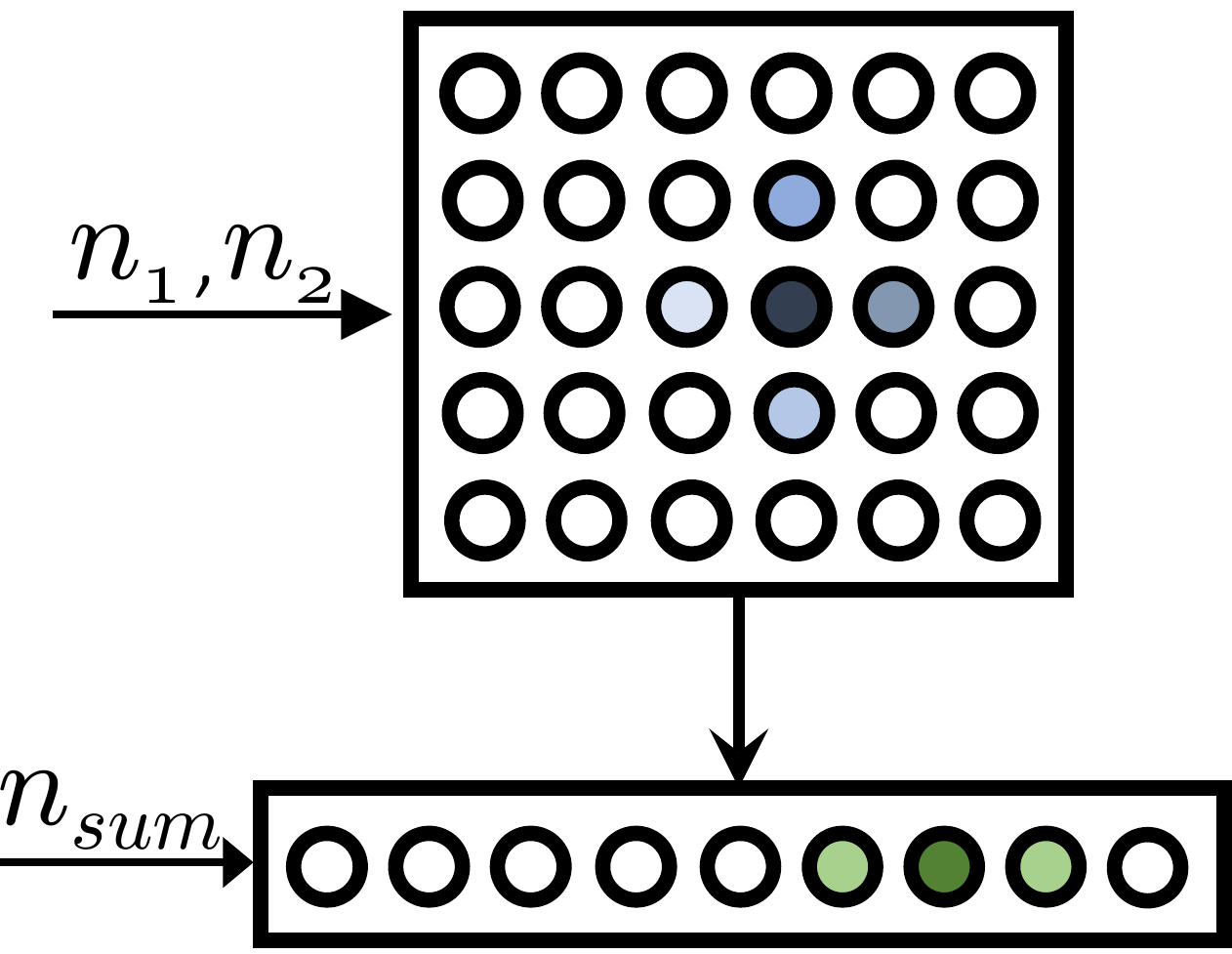}
\caption{Joint encoding}
\label{fig:sum2d}
\end{subfigure}
\caption{A schematic of the two possible ways to perform summation of two variables ($n_1$ and $n_2$). In this paper, (a) is the chosen approach.}
\label{fig:summation}
\end{figure}

To test the proposed network and tuning method, the summation of two variables ($n_1+n_2=n_{sum}$) is carried out in different approaches. 
In Fig. \ref{fig:sum1d}, the sum is estimated using two 1D-layers that encode the two variables (``$n_1$'' and ``$n_2$'') connected through all plastic connections to a 1D-layer encoding the result (``$n_{sum}$''). 
Random numbers are generated within a range, and introduced to the input layers, with their summation introduced into the output layer. 
The network parameters are tuned as described in Sec. \ref{subsection:tuning_params}, such that during the training phase firings in both input and output layers allow the plastic connections to represent the desired summation function. 
After training, the sum is estimated from the decoded values at the output layer. The approach gives a mean error of 7.5\%.

Fig. \ref{fig:sum2d} shows another network layout used to test our method. 
A 2D input layer is used instead of two separate 1D layers. 
The difference from the previous layout is that the neurons' activity is estimated by multiplying the normally distributed activity of both variables. 
The mean error for this layout is around 2\%.
To build the proposed DMSNN, we use the former approach. 
The rationale behind this choice is elaborated in Sec. \ref{sec:discussion}.

\begin{figure}[!b]
\centering
\includegraphics[width=\linewidth]{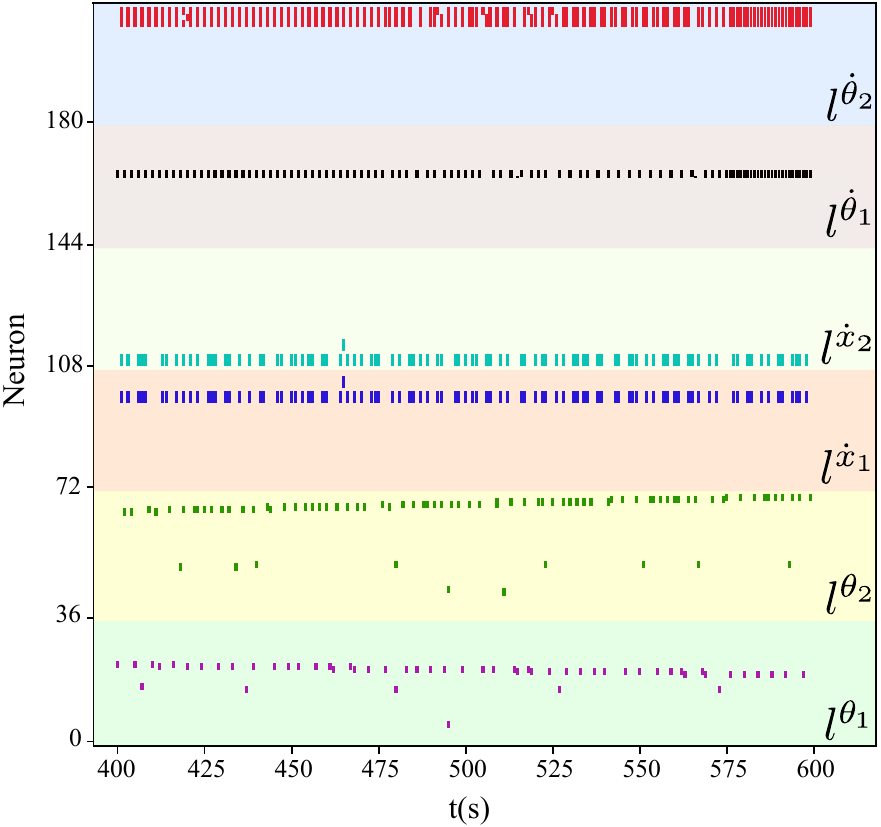} 
\caption{A spike raster plot of neurons in all bundles in the network during the training phase.}
\label{fig:firing}
\end{figure}

\subsection{2DOF Planar Robot}
To verify the effectiveness of our method for visual servoing tasks, a simulation model of a 2DOF planar robot with revolute joints is built. 
The differential kinematics of this system are:
\begin{equation}
    \begin{bmatrix}
    \dot{x}_1 \\ \dot{x}_2
    \end{bmatrix}
    =
    \begin{bmatrix}
    -l_{1} \sin({\theta_{1}}) - l_{2} \sin({\theta_{12}}) & -l_{2} \sin({\theta_{12}})\\ 
    l_{1} \cos({\theta_{1}}) + l_{2} \cos({\theta_{12}}) & l_{2} \cos({\theta_{12}})
    \end{bmatrix} 
    \begin{bmatrix}
    \dot{\theta}_1 \\ \dot{\theta}_2
    \end{bmatrix}
\end{equation}
where $l_i$ denotes the link length, and $\theta_{12} = \theta_1 + \theta_2$.
With the above definition of the Jacobian matrix, we can derive the inverse differential mapping (based on equation \ref{eq:inv_j}), which is then used to generate training data for the SNN. 
During the training process, normalized spatial velocity vectors are fed at random joint angles to the corresponding bundles in the input layer along with the corresponding angular velocities to the bundles of the output layer. The firing activity of the neurons is monitored, as shown in Fig. \ref{fig:firing}, to make sure the network parameters are correctly set to obtain the desired firing frequencies and patterns of the neurons.
For the accuracy of the robot simulation, around the singular configurations, a small value ($10^{-5}$) is added to the determinant of the Jacobian matrix to avoid the unbounded values of joint velocities.

After training, the network is tested by providing a random target in Cartesian workspace $x_{d}$ for a random initial position $x_{i}$. 
By generating a normalized spatial velocity vector $\Vec{v}_{d}$ from the current position $x$ to reach the target, the estimated angular velocities ($\dot{\theta}_{cmd}$) at each joint can be decoded from the bundles of the output layer. 

To quantify the performance, we calculate the minimum distance $\delta$ from the current position to the line segment $\overline{x_{i}x_{d}}$ as shown in Fig. \ref{fig:servo_2d}. 
The target path is a straight line, thus, the shortest distance at each point is defined by the normal to $\overline{x_{i}x_{d}}$. But in other cases, a more complex path may be required. For a robot moving along the path $C$, divided into $N_{C}$ points, given a reference target path $\Omega$, composed of $N_{\Omega}$ points, the mean error $e_{mean}$ is calculated by averaging the maximum deviation $\delta_{max}$ over $N_{trials}$ by applying Algorithm \ref{alg2}.
The training is done over approximately 3000 iterations and tested over 5 times repetition of servoing to 15 different targets (generated randomly in the workspace), to obtain the number of successful trials (in which the final end-effector error $e_x$ is below the threshold value, that is chosen to be 1mm here) and the standard deviation of the maximum deviation $\delta_{max}$ for approaching chosen targets is given by:
\begin{algorithm}[!b]
    \caption{Calculating Mean Error}\label{alg2}
    \begin{algorithmic}[1]
        \State {$\delta_{max}=0$}
        \For{$i=1$ to $N_{trials}$}
            \State {$e_{j}=0$}
            \For{$j=1$ to $N_{C}$}
                \State {$e_{k}=\lVert C(j) - \Omega(1)\rVert$}
                \For{$k=2$ to $N_{\Omega}$}
                    \State $e_{ij}=\lVert C(j) - \Omega(k)\rVert$
                    \If{$e_{ij} \leq e_{k}$}
                        \State $e_{k}=e_{ij}$
                    \EndIf
                \EndFor
                \If{$e_{k} \geq e_{j}$}
                        \State $e_{j}=e_{k}$
                \EndIf
            \EndFor
            \State Stack error {$\delta_{max}(i) \leftarrow e_{j}$}
        \EndFor
        \State{$e_{mean}=\dfrac{1}{N_{trials}}\sum_{n=1}^{N_{trials}}\delta_{max}(n)$}
    \end{algorithmic}
\end{algorithm}

\begin{equation}
\sigma_{servoing} = \sqrt{\frac{\sum_{n=1}^{N_{trials}}(\delta_{max}(n)-e_{mean})^2}{N_{trials}-1}}  
\end{equation}

\begin{figure}[htbp]
	\centering
	\begin{subfigure}{\columnwidth}
	\centering
	\includegraphics[width=\columnwidth]{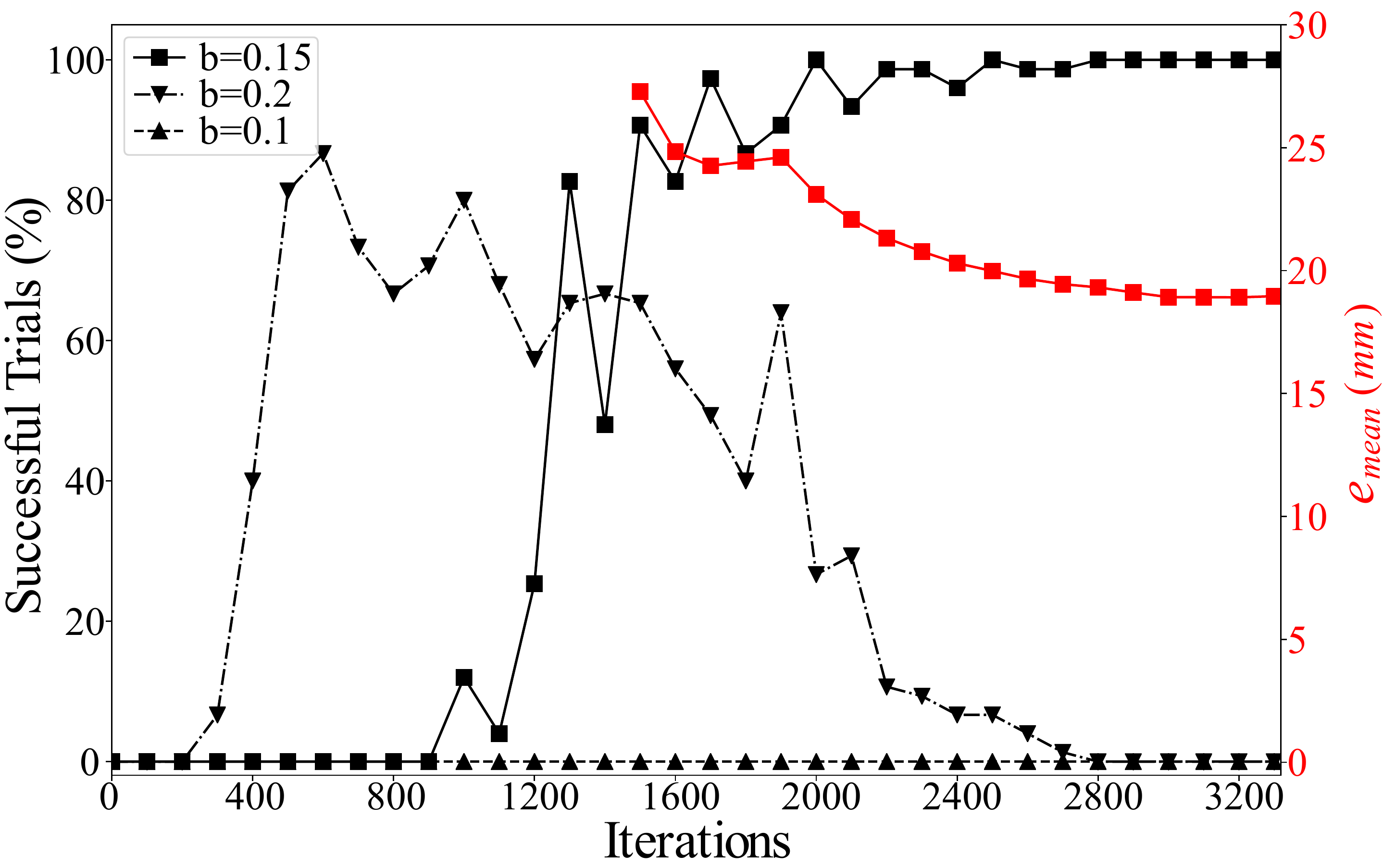}
	\caption{}
	\label{fig:err_trials}
    \end{subfigure}
    \begin{subfigure}{\columnwidth}
	\centering
	\includegraphics[width=\columnwidth]{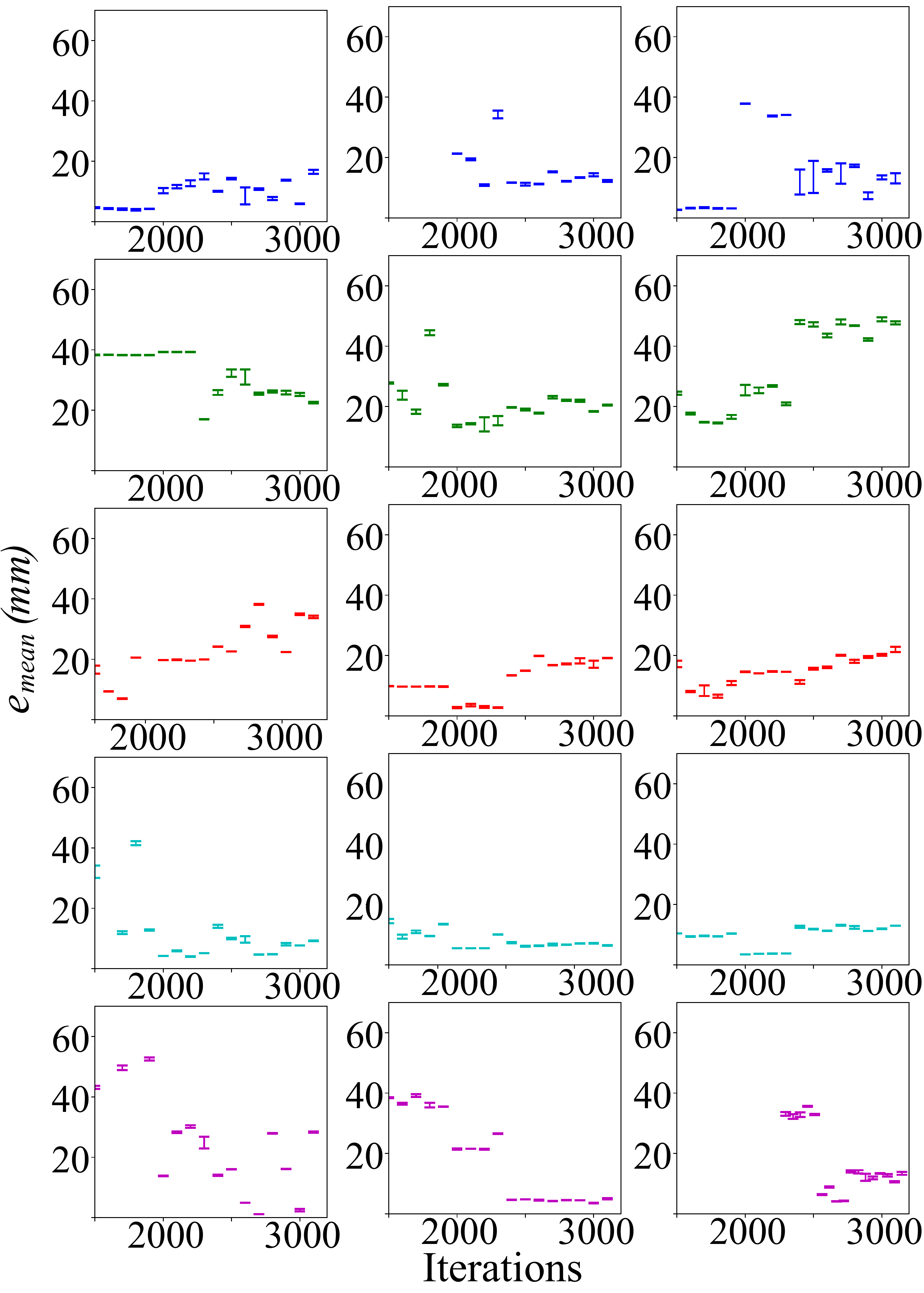}
	\caption{}
	\label{fig:err_std}
    \end{subfigure}
	\caption{(a) Plot of both the number of successful trials (in black), and the mean error (in red) during the control phase versus the number of training iterations. The solid, dotted, and dashed black lines are related to the output neurons with the parameter $b = 0.15, 0.1$ and $0.2$, respectively. The mean error is only plotted for the case in which successful trials are above 80\%. (b) Plot of the mean error and standard deviation for each individual target approach (unsuccessful trials are discarded from calculations and plot).}
\label{fig:err}
\end{figure}

Fig. \ref{fig:err_trials} shows the plot of the percentage of successful trials for the chosen parameters against the number of training iterations. It can be concluded that the learning in that case reaches a stable state after around 2700 iterations. Also, the figure includes the plot of the network performance with a slight change of the key parameters (in this case parameter $b$ for the output neurons as discussed earlier) with a noticeable degradation in the learning ability and stability of the network. Thus, it can be concluded the importance of fine tuning of the network parameters. Additionally, Fig. \ref{fig:err_std} shows the development of the performance of servoing to different points in the workspace. It shall be noted that the performance of the network is degraded in some areas while the training proceeds, thanks to the generalization of the learning process. Thus, instead of training at a local area, the whole defined workspace is instead learnt. This means that initially specific examples are learnt and all the neurons act to represent these examples, but as the training proceeds the representation includes a wider set of examples.

\section{EXPERIMENTS}\label{sec:results}

\subsection{Experimental Setup}
\begin{figure}[!b]
\centering
\begin{subfigure}{0.45\columnwidth}
\centering
\includegraphics[width=0.9\linewidth]{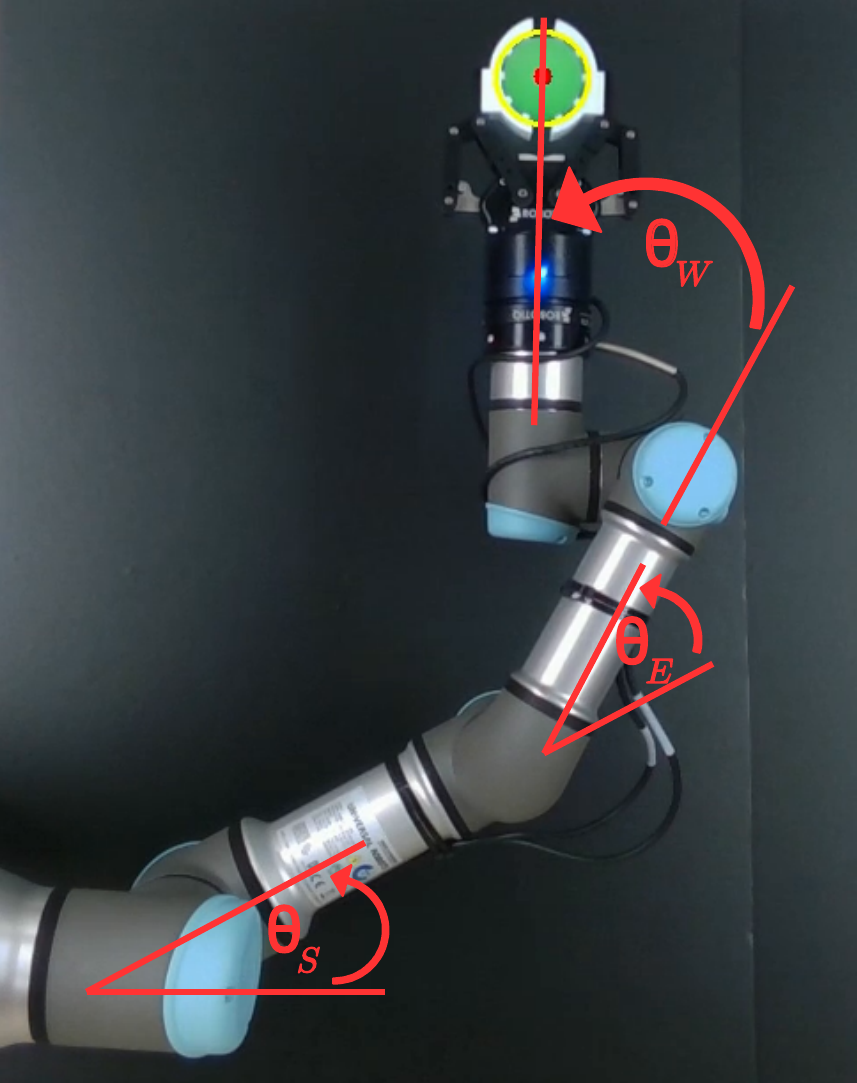} 
\caption{UR3 Planar configuration}
\label{fig:3dof_setup}
\end{subfigure}%
\begin{subfigure}{0.45\columnwidth}
\centering
\includegraphics[width=0.9\linewidth]{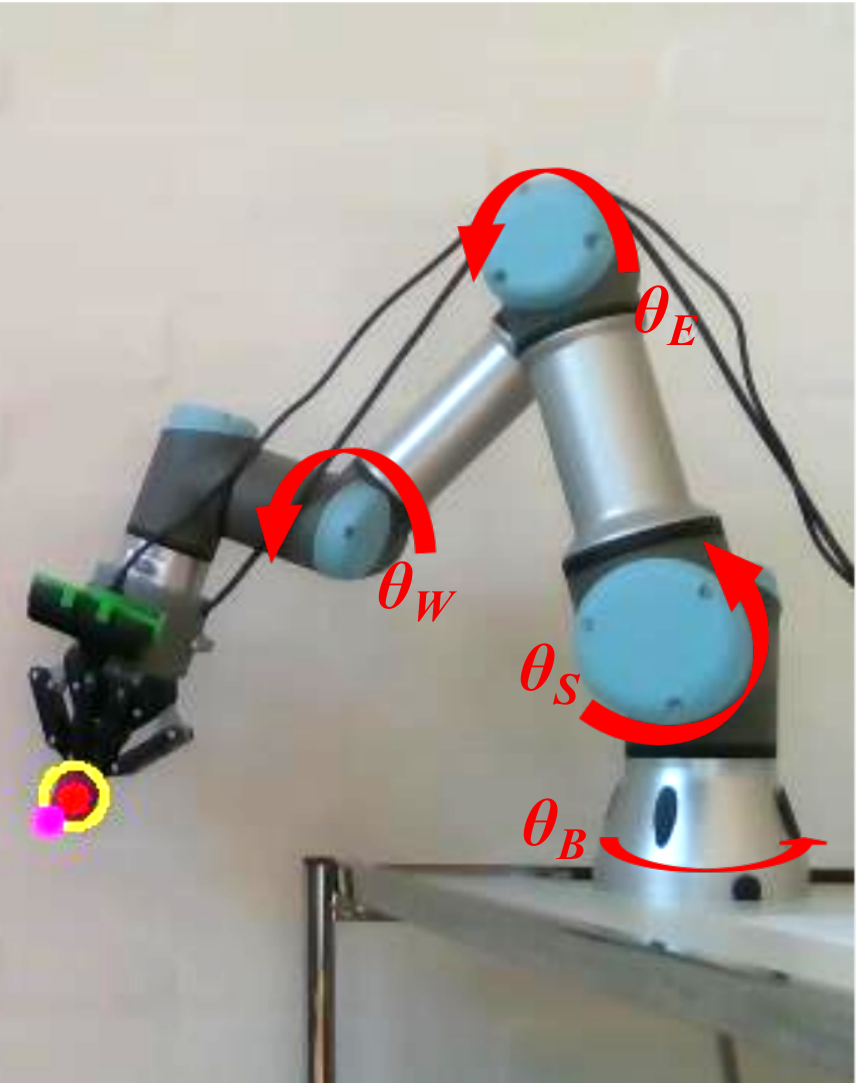}
\caption{UR3 Spatial configuration}
\label{fig:4dof_setup}
\end{subfigure}
\caption{The UR3 moves (a) planar motion while controlling 3DOFs and (b) spatial motion while controlling 4DOFs.}
\label{fig:robo_setup}
\end{figure}

The proposed SNN is simulated using NeMo library, a tool developed to simulate SNN\cite{nemo,gamez2012ispike}. 
A UR3 robot is set with an Intel RealSense D415 camera in setup as shown in Fig. \ref{fig:3dof_setup}. 
The D415 is an RGB-D camera providing real-time color frames and depth maps. 
Image registration is carried out by aligning the color and depth frames, then, the end-effector position can be measured through color filtering of the manipulated object. 
For effective color filtering, the image is blurred to remove high-frequency noise, then the color frame is converted to HSV color space for more robust performance independent of the light intensity. 

A mask, in the desired color range, is applied to obtain a binary image, which is then subjected to some morphology operators to remove the small blobs. 
These operations allow to filter the image and make it easier to discriminate the colored end-effector as the biggest contour. Afterwards, the moment of this contour is calculated such that $M_{ij} = \sigma \phi(x_1,x_2){x_1}^{(i)}{x_2}^{(j)}$.
Then, the centroid ($\rho_x,\rho_y$) is calculated in pixel units based on the moment such that $\rho_x=M_{10}/M_{00}$ and $\rho_y=M_{01}/M_{00}$, see \cite{hu1962visual}. 

The location at the center of the object is then converted to world coordinates by using the camera's intrinsic parameters \cite{Sturm2014}:
\begin{equation}
\label{eqn:eqlabel}
     x_1 = (\rho_{x}-c_{x})\frac{x_3}{F}, \qquad
     x_2 = (\rho_{y}-c_{y})\frac{x_3}{F}
\end{equation}
where $\rho_{xy}$ and $\rho_{y}$ denote to the end-effector position in pixels, $x_3$ is the end-effector's depth, $c_x$ and $c_y$ denote the principal point and $F$ is the focal length.
The following first-order filter is used to remove noise from these visual measurements:
\begin{equation}
    s_{f}(t+1) = s_{f}(t)-\lambda(s_{f}(t)-s(t+1))
\end{equation}
where $s$ and $s_{f}$ are the variable before and after filtering, respectively, and $\lambda>0$ denotes the filter's gain.

\begin{figure}[!t]
\centering
\begin{subfigure}{0.9\columnwidth}
\centering
\includegraphics[width=\linewidth]{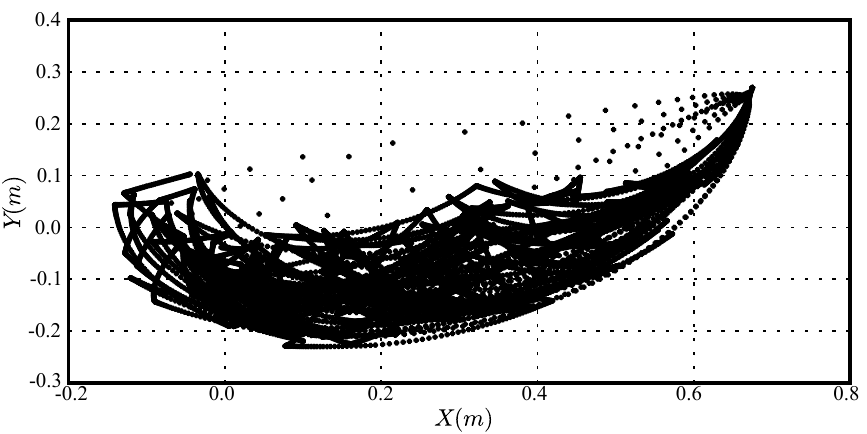} 
\caption{}
\label{fig:train_3dof}
\end{subfigure}
\begin{subfigure}{0.9\columnwidth}
\centering
\includegraphics[width=\linewidth]{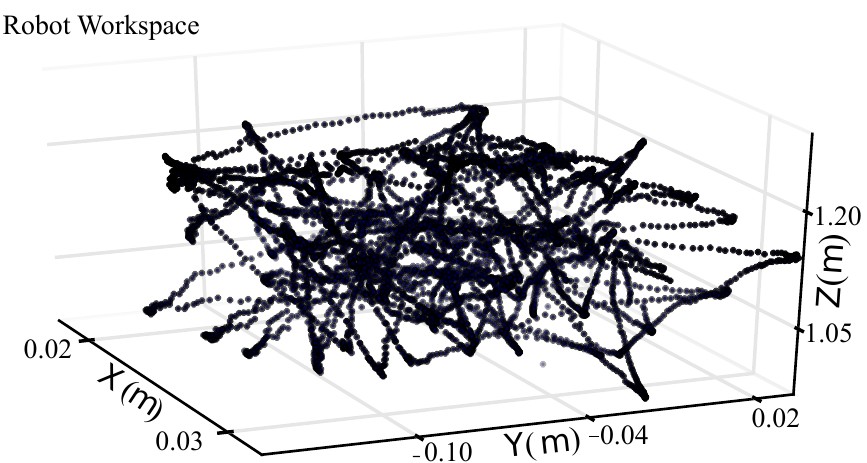} 
\caption{}
\label{fig:train_4dof}
\end{subfigure}
\begin{subfigure}{0.9\columnwidth}
\centering
\includegraphics[width=\linewidth]{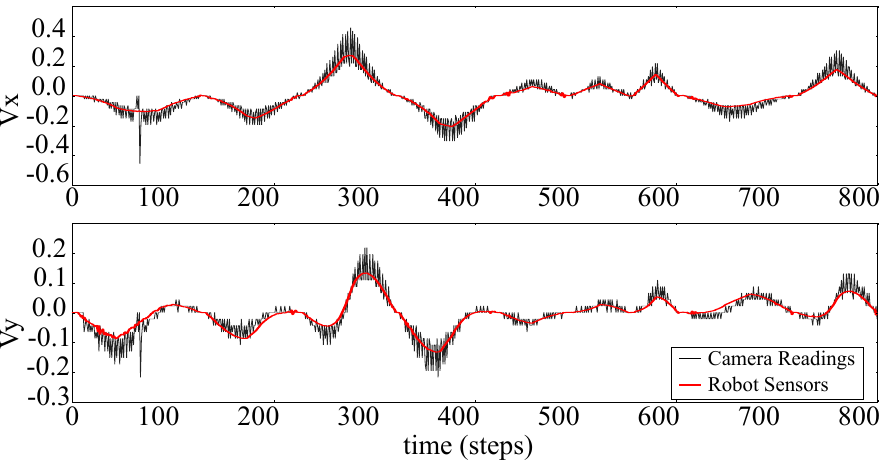}
\caption{}
\label{fig:filter_3dof}
\end{subfigure}
\begin{subfigure}{0.9\columnwidth}
\centering
\includegraphics[width=\linewidth]{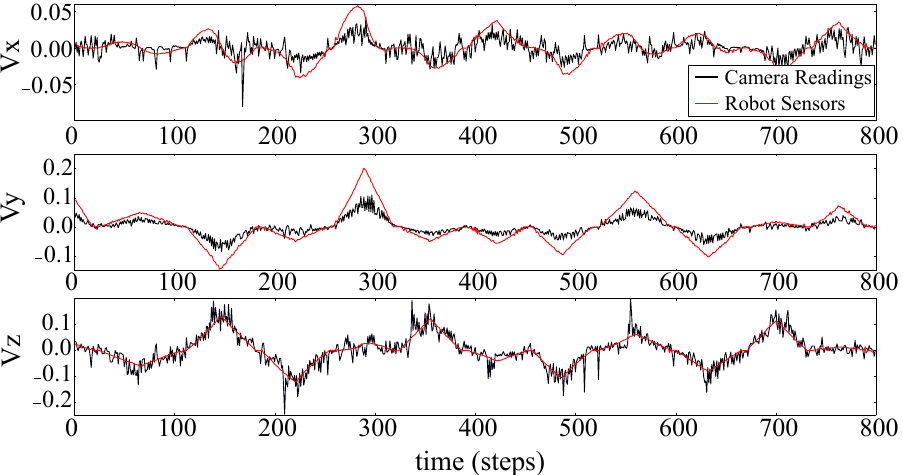}
\caption{}
\label{fig:filter_4dof}
\end{subfigure}
\caption{Plot of data collected for training the DMSNN for (a) 3DOFs and (b) 4DOFs, and velocities as measured from the camera (after filtering) against that estimated by the internal sensors of the robot for (c) 3DOFs and (d) 4DOFs.}
\label{fig:train_data}
\end{figure}

To train the network, the robot is driven linearly in joint space between randomly generated angles. 
Data is collected at a constant sampling rate of 25Hz, then filtered. 
Fig. \ref{fig:filter_3dof} shows the position obtained from the camera before and after filtering. 
Each neuron bundle is fed with the corresponding data to let the plastic connections develop to form the required differential mapping.
Once the training phase ends (dependent upon the number of neurons and size of workspace), the strength of the synaptic connections is kept constant. 
Random targets are given across the robot workspace as shown in Fig. \ref{fig:train_3dof} and Fig. \ref{fig:train_4dof}. 
The current joint angles and the desired spatial velocity are updated every 20ms.

\subsection{3DOF Planar Robot}
\begin{figure}[!t]
  \centering
  \includegraphics[width=\columnwidth]{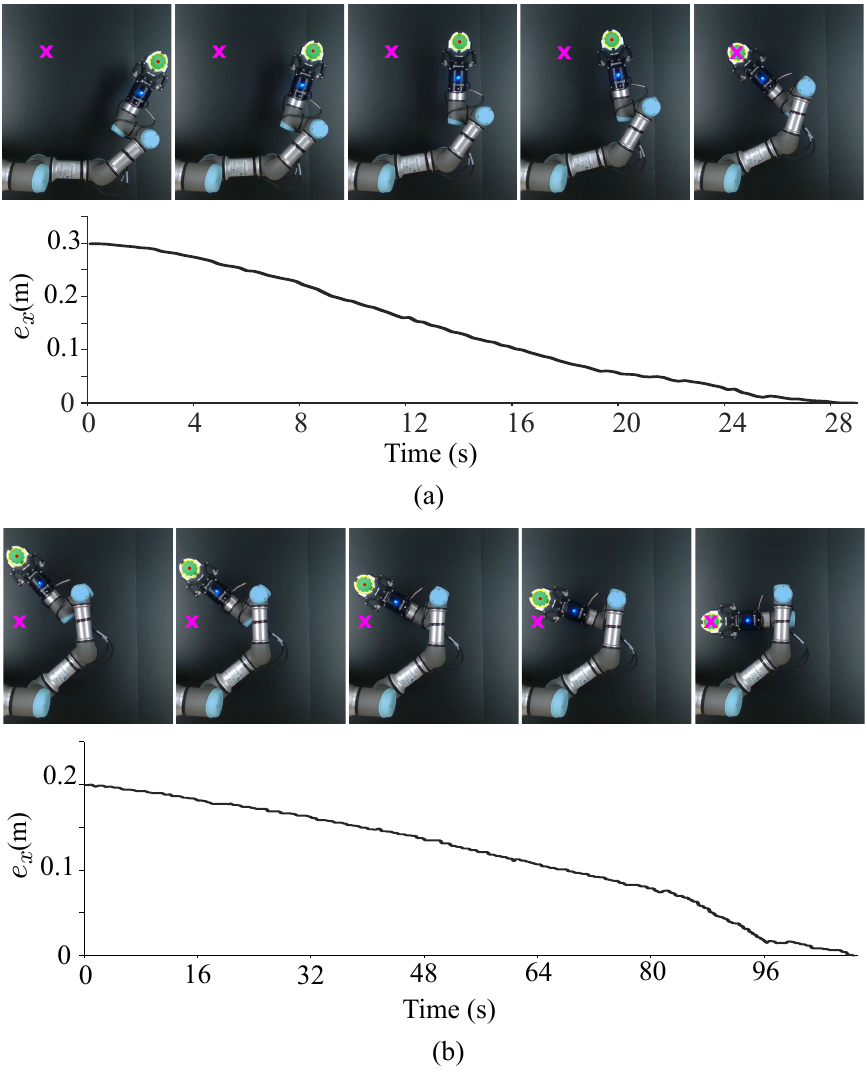}
  \caption{The UR3 while performing planar motions in (a) and (b) with the norm of the error $e_x$ plotted during motion.}
\label{fig:3dof_servoing}
\end{figure}

For this case, three joints are controlled to guide the UR3 robot in planar motion, as shown in Fig. \ref{fig:3dof_servoing}. 
The network consists of five input bundles ($l^{\theta_{1:3}}$ and $l^{\Dot{x}_{1:2}}$) and three output bundles ($l^{\Dot{\theta}_{1:3}}$). 
The network parameters, as well as the number of neurons in each bundle, are shown in Table \ref{table:net_params}. 
These three joints have ranges of: $[-180\degree,-90\degree], [-45\degree,0\degree]$ and $[90\degree,180\degree]$. 

\begin{figure}[!t]
\centering
\includegraphics[width=0.83\columnwidth, height=2.22\linewidth]{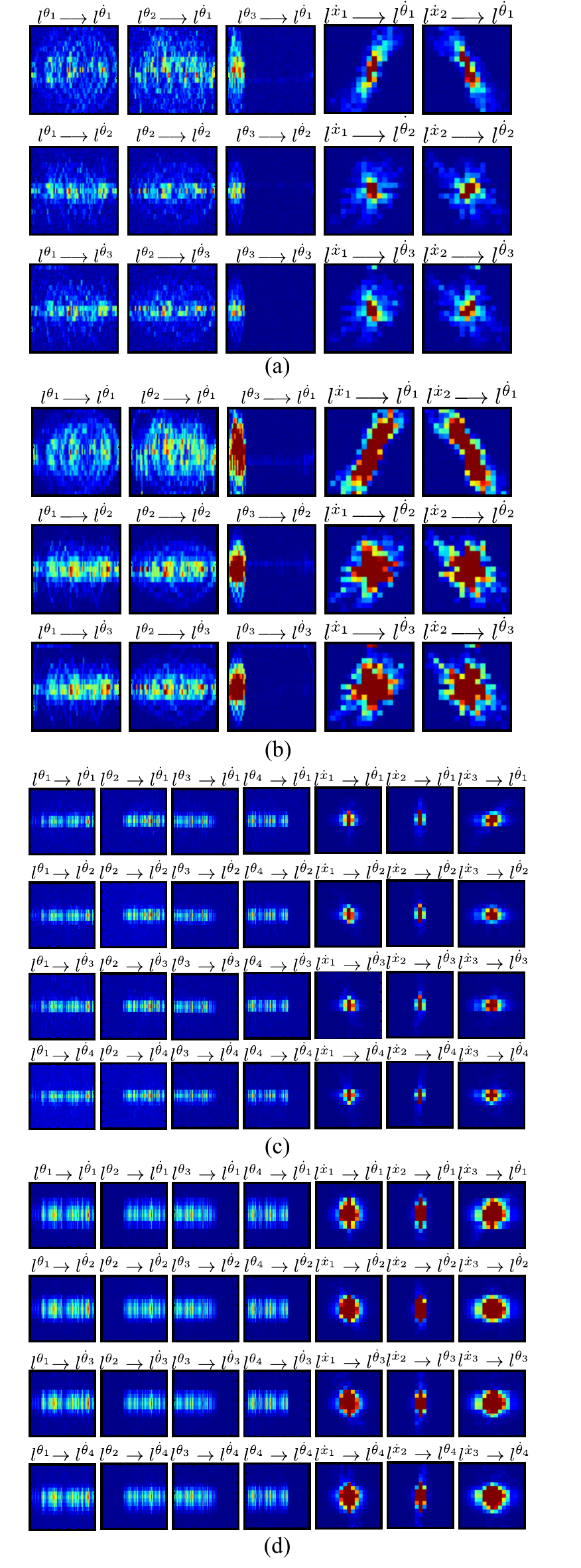} 
\caption{Heatmap of the weights update process for the 3DOF case of study at (a) 3000 and (b) 6000 iterations, and for 4DOF case at (c) 4000 and (d) 9000 iterations. The dark blue color corresponds to zero weight while dark red color corresponds to the maximum weight.}
\label{fig:weights_update_3_4_dof}
\end{figure}

From the motor babbling data collected, the maximum and minimum values for both the spatial and angular values are obtained. 
For each layer with $\mathcal{N}_{3D}$ neurons, the central value $\psi_{c}$ encoded by each neuron is assigned by dividing the range of each variable evenly over the whole layer.
The update of the synaptic strength is shown as heatmaps in Fig. \ref{fig:weights_update_3_4_dof}a and Fig. \ref{fig:weights_update_3_4_dof}b. 
Each heatmap depicts the relation between one bundle from the sensory layer to one bundle from the motor layer, where each pixel gives the strength of an excitatory synapse connecting one sensory neuron to one motor neuron.
After running the simulation for 6000 iterations, the strength of synapses between a motor layer to the sensory layer is modulated to represent the required differential map. 
The robot is then given random points to reach by its end-effector through a visual servoing process, and considered successful at reaching if the end-effector is less than 3 pixel away from the target in each coordinate. As shown in Fig. \ref{fig:3dof_servoing}, a target point is provided and the robot automatically moves towards it by using decoded motion command $\Dot{\theta}_{cmd}$. 
The distance between the end-effector and the target is represented as the norm $\|e_x\| = \|x - x_d \|$.\\

\begin{table}[!b]
\caption{Network Parameters}
\begin{tabular}{|c||*{5}{c|}}
 \hline
 \multicolumn{6}{|c|}{Neuronal layers} \\
 \hline
 \hline
\backslashbox[20mm]{Layer}{Param}&$a$&$b$&$c$&$d$&$\mathcal{N}_{3D/4D}$\\ \hline
$l^{\theta_{i}}$&0.1&0.2&-65&2&68/136\\
$l^{\Dot{x_{i}}}$&0.1&0.2&-65&2&68/136\\
$l^{\Dot{\theta_{i}}}$&0.02&0.15&-55&6&68/136\\
 \hline
 \end{tabular}
 \begin{tabular}{|c||*{4}{c|}}
 \hline
 \multicolumn{5}{|c|}{Synaptic connections} \\
 \hline
 \hline
 \backslashbox[20mm]{DOF}{Param}& $S$ & $\tau_{1}/\tau_{2}$ & $C_{I}/C_{E}$& $Itr^{*}$\\
 \hline
 3 (Planar) & 0.05 & 20/18 & -4/4 & 6000\\
 4 (Spatial) & 0.03 & 20/18 & -5/5 & 9000\\
 \hline
\end{tabular}
\label{table:net_params}
\end{table}

\subsection{4-DOF Spatial Robot}
In this case, four joints are controlled to guide the UR3 robot as shown in Fig. \ref{fig:4dof_setup}. 

\begin{figure}[!t]
  \centering
  \includegraphics[width=\columnwidth]{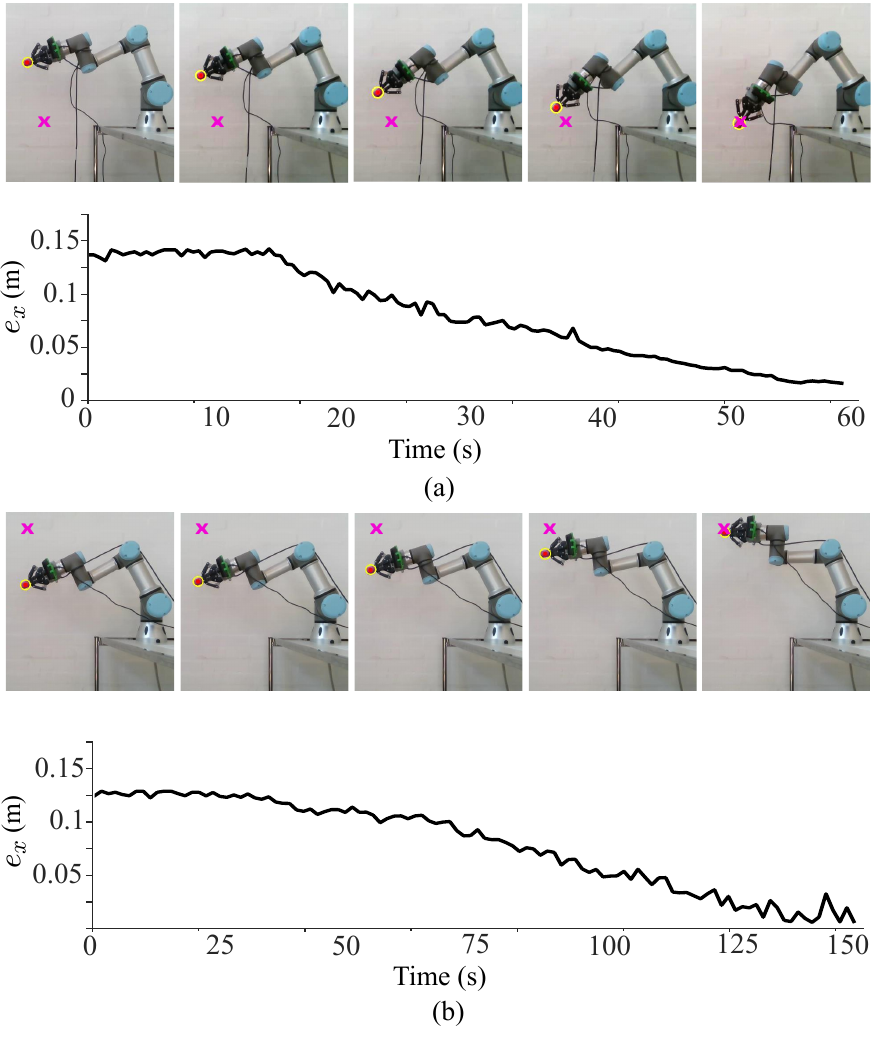}
  \caption{The UR3 while performing spatial motions in (a) and (b) with the norm of the error in the 3 axes $e_x$ plotted during motion.}
\label{fig:4dof_servoing}
\end{figure}
The network consists of seven input groups ($l^{\theta_{1:4}}$ and $l^{\Dot{x}_{1:3}}$) and four output groups ($l^{\Dot{\theta}_{1:4}}$).
The network parameters and the number of neurons in each layer are shown in table \ref{table:net_params}.
The four joints ranges are: $[-200\degree,-170\degree], [-75\degree,-45\degree], [90\degree,110\degree]$ and $[-200\degree,-160\degree]$. 
The heatmaps depicting the update of weights at the 4000 and 9000 iterations are shown in Fig. \ref{fig:weights_update_3_4_dof}c and Fig. \ref{fig:weights_update_3_4_dof}d, respectively. 
Similar to the previous setup, several end-effector targets (randomly given across the work space)  are given to the robot in sensor space. The target reaching in this case is considered successful when the end-effector is less than 3 pixel away from the target in $x$ and $y$ coordinates and 5mm in depth ($z$ coordinate).
Representative results for these 4-DOF visual servoing tasks are depicted in Fig. \ref{fig:4dof_servoing}\textcolor{rephrase}{, and it shall be noted that the ripples in the plot are due to the noise in the data collected by the depth sensor, which can be judged by comparison to the plots of velocities based on the readings collected by the motor encoders as shown in Fig. \ref{fig:filter_4dof}. }
The accompanying multimedia video demonstrates the performance of our new method with many experimental results.
\FloatBarrier

\section{DISCUSSION}\label{sec:discussion}
As shown in section \ref{subsection:summation}, the accuracy of the summation of two numbers by the network in the case of multiple one-dimensional arrays is lower than that obtained in the case of a multi-dimensional array. 
However, this result comes on the expense of the network size where the 1D layers require only 2$\mathcal{N}$ neurons, for $\mathcal{N}$ is the number of neurons per layer, while the 2D layer needs $\mathcal{N}^{2}$ neurons. 
This means that for an $O$-dimensional case, $\mathcal{N}^{O}$ neurons would be needed. 
However, note that the explicit use of visual feedback in our formulation provides a valuable \emph{rectifying} property to the network (i.e. it corrects for small errors in an otherwise open-loop controller) while demanding moderate computational power.

Moreover, in this study the SNN is used to guide the robot for visual servoing process, where it requires fast state updates, hence the proposed network is to be used instead as it compromises between real-time operation and sufficient accuracy. 
However, in case of accurate mapping of robot states and commands the second approach is to be used, or additional hidden layers are to be added to the network for higher accuracy and better estimates. 
Another feature, which is not evident from the summation test studied, is that SNN better approximates real-life operations where the values to be encoded/decoded are continuous, i.e. with no discontinuity or jumps. 
For the neurons in the network to evoke a spike, it needs first to let the membrane potential build up over a series of time steps. While incorporating the Gaussian activity, this leads to sharing excitation with neurons in the neighborhood, and it becomes more likely for the adjacent neurons to be triggered in next cycles as well. As the adjacent neurons encode values close to each other, this achieves a continuous smooth operation and avoid sudden changes and jerky motions. \textcolor{rephrase}{So, despite the fact that in the asynchronous SNN, as in our case, neurons that release a spike first tend to spike again, the fine tuning of the parameters acts to facilitate more accurate approximations and the neighboring neurons activity acts to correct the estimations during the servoing process.}

This work also exploits the potential of an SNN based on STDP, following some procedures for fine-tuning of the network parameters as described in subsection \ref{subsection:tuning_params}, where only two layers (input and output) and a low number of neurons are needed to develop a differential map for multi-DOF robot. \textcolor{rephrase}{Additionally, the noisy data is well handled thanks to the inhibition that occurs for distal firing of pre and post-synaptic neurons, which weakens the synapses connecting uncorrelated neurons, aside to the inter-inhibitory connections that inhibits distal neurons to avoid undesired firing.}

\textcolor{rephrase}{ The developed network can then be compared to the work from \cite{wu20082d}, where a SNN is used to build a map for static transformation of coordinates for a 2DOF robot (let us denote it as \textit{CTSNN}), and the work from \cite{neymotin2013reinforcement}, where an SNN modeling the sensorimotor cortex is used to guide a 2DOF robot in a reaching task (and denoted as \textit{SMCSNN}). The comparison is concluded in Table \ref{table:compare} to show the time and number of iterations needed for training the networks, the network  size, and the final error $e_x$.}
\begin{table}[!b]
\caption{Networks Comparison}
\begin{threeparttable}
\begin{tabular}{|c||p{10mm}|p{7mm}|p{10mm}|p{9mm}|}
 \hline
\backslashbox[19mm]{Net}{Param}&$Itr^*$&$T^*(s)$&$e_x$&$\mathcal{N}_{net}$\\ \hline
\textit{CTSNN}&$\approx$4000&800&NA&2000\\
\textit{SMCSNN}&200$ep$&3000&$\approx$1mm&704\\
\textit{DMSNN}&3000&30&$\leq$1mm&216\\
 \hline
\end{tabular}
\begin{tablenotes}
   \item[*] The number of iterations for \textit{CTSNN} is estimated based on 8000s training divided into 1600 time steps (each time step 0.125ms) for each stimulus introduced. The number of training steps is not mentioned for the \textit{SMCSNN}, instead the number of epochs is mentioned (which in this case is the motion from one starting point to a target point) and each epoch takes around 15s. Each iteration in the DMSNN is 80ms of simulation time (which is only 10ms of real time), through which a certain stimulus is introduced to the network.
\end{tablenotes}
\end{threeparttable}
\label{table:compare}
\end{table}

\textcolor{rephrase}{It can be concluded that both the simulation time and the real time required for training and running the network is less for the proposed network, thanks to the fine tuning of the network parameters and maximizing the benefit from the spiking nature of the network.} 

Following the proposed tuning method, each neuron bundle consists of around 136 neurons, in case of the spatial configuration, compared to around 1000 neurons in \cite{murray}.
In \cite{hbp_hand} each bundle is constructed of around 100 neurons only, however, a hidden layer is needed which increases the size of the network and complexity of the tuning process. 
Moreover, the robot in this work needed to approach only 100 points, and data recorded while moving to these target points was sufficient for the training process compared to 6426 target points in \cite{tieck2018controlling}.
The robot succeeded in all trials to reach the destined targets through visual servoing.

The time required to reach the target varies depending on the distance from the current position and the richness of data trials available along the path between the current and target pose. This can be seen in the two examples in Fig. \ref{fig:3dof_servoing}, where in Fig. \ref{fig:3dof_servoing}a the robot takes less time to cover a bigger distance compared to Fig. \ref{fig:3dof_servoing}b. This can be explained by having the training data collected from motor babbling in joint space with more examples of the robot moving in waving-like trajectories and less abundant data for linear motion in task space. 
Similarly, the trial in Fig. \ref{fig:4dof_servoing}a less time is needed compared to the trial in Fig.\ref{fig:4dof_servoing}b as the target in the latter is close to the boundary of the studied section of the workspace which has less training examples. 
Richer data collected from motions in both joint space and task space can be used in a future study to role out the effect of varying the data on different motion types, and the possibility of emerging of motion primitives from such behavior.

It can be noticed, from Fig. \ref{fig:filter_3dof} and Fig. \ref{fig:filter_4dof}, the training data is filtered using only first-order filter and contains more noise in case of the spatial motion, due to the noise in depth readings, however, the DMSNN is still able to approximate the differential relationship and build the desired map. This can be justified by the depression in the strength of uncorrelated neurons due to the chosen STDP rule.

\section{CONCLUSIONS}\label{sec:conclusions}
\textcolor{rephrase}{Based on an intuitive architecture and tuning guideline,} the proposed DMSNN provides a way to build a differential map to drive robots with many DOFs through multi-modal servoing tasks.
The network is featured by real-time operation and a small amount of data, compared to similar methods in literature, is needed for training.
The current limitations of this method are related to the limited resolution of the network output.
Future work will focus on improving the output resolution, as well as deriving a mathematical formula to conclude the optimal parameters for neurons firing and STDP learning.
We also plan to use this method for conducting vision-guided shape control tasks with deformable objects, as in \cite{dna2018_tro}, and apply a cerebellar controller to enhance the performance and reduce the deviation from the path, as in \cite{corchado2019integration}.

\section*{Funding}
This research work is supported in part by the Research Grants Council (RGC) of Hong Kong under grant number 14203917, in part by PROCORE-France/Hong Kong Joint Research Scheme sponsored by the RGC and the Consulate General of France in Hong Kong under grant F-PolyU503/18, in part by the Chinese National Engineering Research Centre for Steel Construction (Hong Kong Branch) at PolyU under grant BBV8, in part by the Key-Area Research and Development Program of Guangdong Province 2020 under project 76 and in part by The Hong Kong Polytechnic University under grant G-YBYT and 4-ZZHJ.

\bibliography{biblio.bib}
\bibliographystyle{IEEEtran}

\end{document}